\pdfoutput=1
\documentclass[11pt]{article}
\usepackage[]{acl}
\usepackage{times}
\usepackage{latexsym}
\usepackage[T1]{fontenc}
\usepackage[utf8]{inputenc}
\usepackage{microtype}
\usepackage{inconsolata}
\usepackage{amsmath}
\usepackage{float}
\usepackage{amssymb}
\usepackage{xspace}
\usepackage{booktabs}
\usepackage{multirow}
\usepackage{graphicx}
\usepackage{tabularx}
\usepackage{subcaption}
\usepackage{tcolorbox}

\newcommand{\paratitle}[1]{\vspace{1.5ex}\noindent\textbf{#1}}
\newcommand{\ie}{\emph{i.e.,}\xspace}

\newcommand{\eg}{\emph{e.g.,}\xspace}

\newcommand{\ignore}[1]{}

\newcommand{\ICL}{$_\texttt{ICL}$}
\newcommand{\TR}{$_\texttt{TR}$}
\newcommand{\TL}{$_\texttt{TL}$}

\title{Investigating the Pre-Training Dynamics of In-Context Learning: \\Task Recognition vs. Task Learning}

\author{
    \textbf{
        Xiaolei Wang\textsuperscript{\rm{1,3}\thanks{\ \ Equal contribution.}},
        Xinyu Tang\textsuperscript{\rm{1,3}\footnotemark[1]},
        Wayne Xin Zhao\textsuperscript{\rm{1,3}\thanks{\ \ Corresponding author.}}, 
        Ji-Rong Wen\textsuperscript{\rm{1,2,3}}
    }\\
    \textsuperscript{1}Gaoling School of Artificial Intelligence, Renmin University of China\\
    \textsuperscript{2}School of Information, Renmin University of China\\
    \textsuperscript{3}Beijing Key Laboratory of Big Data Management and Analysis Methods\\
    \texttt{wxl1999@foxmail.com, txy20010310@163.com, batmanfly@gmail.com}\\
}

\begin{document}
\maketitle

\begin{abstract}

The emergence of in-context learning~(ICL) is potentially attributed to two major abilities: \textit{task recognition}~(TR) for recognizing the task from demonstrations and utilizing pre-trained priors, and \textit{task learning}~(TL) for learning from demonstrations.
However, relationships between the two abilities and how such relationships affect the emergence of ICL is unclear.
In this paper, we take the first step by examining the pre-training dynamics of the emergence of ICL.
With carefully designed metrics, we find that these two abilities are, in fact, \textit{competitive} during pre-training.
Moreover, we observe a strong negative correlation between the competition and ICL performance.
Further analysis of common pre-training factors (\ie model size, dataset size, and data curriculum) demonstrates possible ways to regulate the competition.
Based on these insights, we propose a simple yet effective method to better integrate these two abilities for ICL at inference time.
Through adaptive ensemble learning, the performance of ICL can be significantly boosted, enabling two small models to outperform a larger one with more than twice the parameters.
The code is available at \url{https://github.com/RUCAIBox/Competitive-ICL}.

\end{abstract}
\section{Introduction}
\label{sec-introdction}

In-context learning~(ICL)~\cite{brown2020language} represents a significant advancement in the capabilities of large language models~(LLMs).
It allows models to rapidly adapt to new tasks without updating the parameters by adding only a few examples as demonstrations to the input.
This capability has profound applications on a wide range of tasks~\cite{dong2022survey, lin2023unlocking}.
%Despite this, the underlying mechanism of ICL is still under-explored.

To explore the underlying mechanism, existing work~\cite{Disentangle-ACL-2023, wei2023larger} mainly focuses on how LLMs perform ICL during inference.
Two main abilities are considered to play important roles in ICL: task recognition~(TR), which recognizes the task from demonstrations and utilizes pre-trained priors, and task learning~(TL), which directly learns to solve the task from demonstrations.
Furthermore, recent research~\cite{Disentangle-ACL-2023} has found that TR is relatively easier to obtain and can be observed in small models with only 350M parameters, while TL would often emerge in large models with billions of parameters.
Based on this, \citet{wei2023larger} further explore the relationships between these two abilities and show that TR takes the dominant in smaller LLMs while TL is more emphasized in larger LLMs.
However, how these two abilities quantitatively affect the emergence of ICL is under-explored.

% \begin{figure}[t]
%     \centering
%     \includegraphics[width=\columnwidth]{figures/MiniCPM-Performance.pdf}
%     \caption{
%         The performance of MiniCPM-2B for ICL and its two abilities (\ie task recognition and task learning). The emergence of ICL encounters many fluctuations, where the performance of task recognition and task learning changes in the opposite direction.
%     }
% \label{fig:performance-minicpm}
% \end{figure}
\begin{figure}[t]
    \centering
    \begin{subfigure}[b]{0.49\linewidth}
        \centering
        \includegraphics[width=\linewidth]{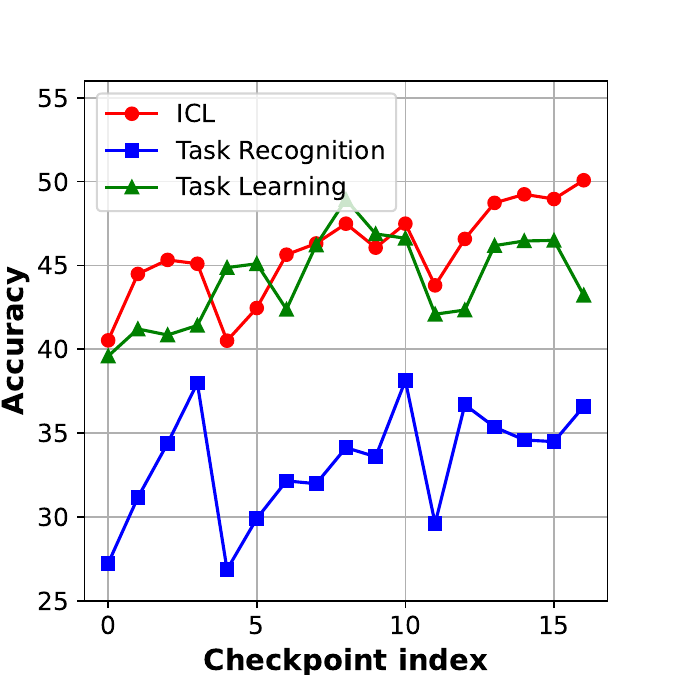}
        \caption{MiniCPM-2B}
        \label{fig:minicpm-preformance}
    \end{subfigure}
    \begin{subfigure}[b]{0.49\linewidth}
        \centering
        \includegraphics[width=\linewidth]{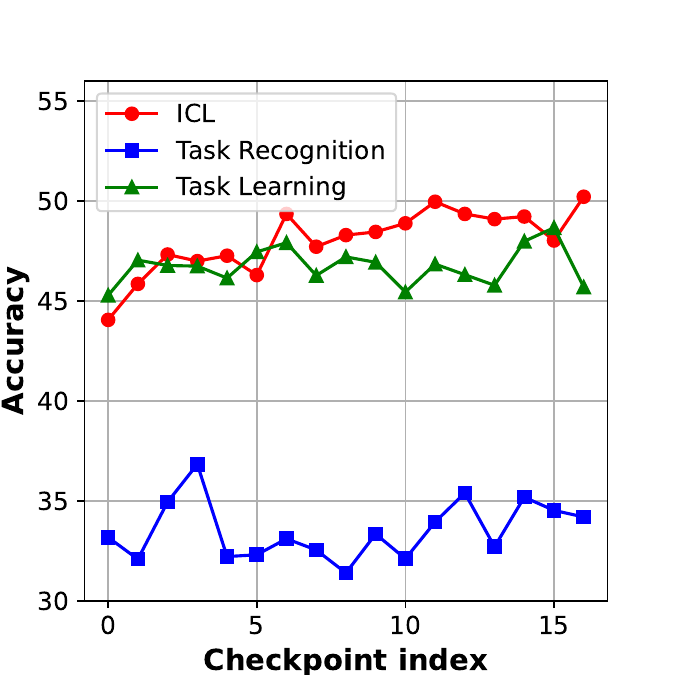}
        \caption{Amber-7B}
        \label{fig:amber-preformance}
    \end{subfigure}
    % \begin{subfigure}[b]{0.49\linewidth}
    %     \centering
    %     \includegraphics[width=\textwidth]{figures/OLMo-Performance.pdf}
    %     \caption{OLMo-7B}
    %     \label{fig:olmo-preformance}
    % \end{subfigure}
    \caption{The performance of MiniCPM-2B and Amber-7B for ICL and its two abilities (\ie task recognition and task learning). The emergence of ICL encounters many fluctuations, where the performance of task recognition and task learning changes in the opposite direction.}
\label{fig:performance-three-setting}
\end{figure}

In this work, we take the first step towards unraveling the mystery, \ie \emph{relationships between the two abilities and how such relationships affect the emergence of ICL}, by examining the pre-training dynamics of LLMs.
To achieve this goal, we first propose to disentangle the two abilities by manipulating the input-label settings~\cite{Disentangle-ACL-2023}, so as to measure the performance of TR and TL for each model checkpoint in the pre-training.  
As illustrated in \figurename~\ref{fig:performance-three-setting}, we can observe that the emergence of ICL encounters many fluctuations, along with competition between its two abilities (\ie their performance actually changes in the opposite direction).
To quantify such a competitive relationship, we propose new measurements to reflect how one ability suppresses the other.

With the proposed metrics, we find that the competitive relationship widely exists for existing LLMs with various training settings.
%is a widespread phenomenon, which can be observed in LLMs with various training settings.
More importantly, it demonstrates a strong correlation with ICL.
First, during pre-training, the competition exhibits a ``stable--rise'' pattern, typically reflecting fluctuations and improvements in the performance of ICL.
Second, with respect to the entire pre-training process, the average intensity of competition (defined in Section~\ref{sec-background}) is negatively correlated with the final ICL performance: the less competition, the better the ICL performance.
These findings suggest that regulating the competition between the two abilities of ICL could be crucial for its emergence. 
We further investigate the influence of common pre-training factors (\ie model size, dataset size, and data curriculum) on the competition.
We find that: (1) scaling model size can lead to the early appearance of competition but effectively reduce the average intensity of competition;
(2) scaling dataset size can postpone the competition\ignore{ and also reduce the average intensity of competition};
and (3) specific data curricula can adjust the intensity of competition for the enhancement or specialization of LLMs.

Furthermore, we propose a simple yet effective method to fuse the two abilities of ICL for better performance at inference time.
Specifically, we first select two checkpoints from the pre-training process with the best abilities of TR and TL, respectively.
Then, they are fused with adaptive ensemble learning, where the contribution of each one is adaptively determined by its performance.
To validate the effectiveness of our approach, we conduct experiments on extensive datasets and LLMs with various training settings.
Experimental results show that this simple method can effectively boost the performance of ICL and outperform several competitive baselines, even if the total parameters are less than half of a single larger LLM.

Our contributions can be summarized as follows:

$\bullet$ To the best of our knowledge, this is the first time that the competitive relationship between the two abilities of ICL (\ie TR and TL) and its emergence has been investigated.
By examining the pre-training dynamics of ICL, we demonstrate a strong negative correlation between the emergence of ICL and the competition between TR and TL.

$\bullet$ We conduct a fine-grained analysis of common pre-training factors (\ie model size, dataset size, and data curriculum) to understand their influence on the competition between TR and TL.

$\bullet$ We propose a simple but effective method to better integrate TR and TL for ICL at inference time.
Through adaptive ensemble learning, the performance of ICL can be significantly boosted, enabling two small models to outperform a larger one with more than twice the parameters.

\section{Background and Measurement}
\label{sec-background}

In this section, we introduce the background of TR and TL and further propose new measurements to quantify the competition between them. 
%two abilities in ICL, metrics to measure their competition, and the experimental setup used in this work.

\subsection{Task Recognition and Task Learning}

Typically, an LLM performs ICL by using input-label pairs from the target task as demonstrations, \ie $D_k = \{ (x_1, y_1), \dots, (x_k, y_k) \}$, to predict the label for the test input. 
In {existing literature~\cite{Disentangle-ACL-2023, lin2024dual}}, it has been widely recognized that ICL can be attributed to two major underlying abilities, namely \emph{task recognition~(TR)} and \emph{task learning~(TL)}.
% We call this the \textit{gold} setting for evaluating the ICL ability.
Specifically, TR refers to the ability of an LLM to recognize the target task from demonstrations and only utilize its own knowledge obtained from pre-training to solve the task, while TL refers to the ability of an LLM to solve the target task solely based on demonstrations.

To disentangle ICL into the two main abilities, existing studies~\cite{Disentangle-ACL-2023, lin2024dual} are mainly developed based on an important assumption: the mapping information between input and label in demonstrations is more important for TL.
%Compared with TR, the mapping information in demonstrations is more important for TL.
Following \citet{Disentangle-ACL-2023}, we consider three settings to study the effect of TR and TL: 

%Existing work~\cite{Disentangle-ACL-2023} empirically disentangles ICL into two main abilities: \textit{task recognition} and \textit{task learning}.

\ignore{Given a set of input-label pair demonstrations with $k$ examples $D_k = \{ f(x_1, y_1), \dots, f(x_k, y_k) \}$ and a new input query $x_{k+1}$, LLMs perform ICL to generate the output $\hat{y}_{k+1}$.
The essence of ICL is to enable the LLMs to learn the mapping $f: \mathcal{X} \to \mathcal{Y}, x_i \in \mathcal{X}, y_i \in \mathcal{Y}$ through the examples.
However, the standard ICL paradigm uses golden input-label pairs, reflecting both task recognition and task learning abilities.
To conduct a fine-grained exploration, we study the disentangled dual abilities in the following section.}

$\bullet$~\emph{Gold}:
It refers to the standard ICL setting, where we use the correct input-label pairs.
This reflects both TR and TL abilities.

% \paratitle{Task Recognition.}

% That means TR does not rely on the mapping information between inputs and labels in the demonstrations.
$\bullet$~\emph{Random}:
To evaluate TR ability, we randomly sample labels from the label space of the target task for each input in demonstrations.
% Such an evaluation method is usually called the \textit{random} setting.

\ignore{Task recognition~(TR) refers to utilizing the pre-trained prior of LLMs to learn the mapping $f$.
It only observe the input distributions ${\{x_i\}_{i=1}^k}$ and label space ${\{y_i\}_{i=1}^k}$ without relying on the mapping $x_i \to y_i$.
Following ~\citet{Rethinking-2022-EMNLP}, when evaluating TR ability, demonstrations are formed with random labels instead of golden labels.
Specifically, each $x_i$ is paired with $\widetilde{y_i}$, which are randomly sampled from the label space ${\{y_i\}_{i=1}^k}$.}

% 任务识别指的是利用模型的先验学习映射 $f$。
% 具体来说，任务识别仅通过捕捉input的分布 ${\{x_i\}_{i=1}^k}$ 和 label的分布 ${\{y_i\}_{i=1}^k}$ ，而不依赖 $x_i \to y_i$ 的映射。
% Following ~\citet{Rethinking}, 在评测任务识别能力时， demonstrations are formed with random labels 而不是 golden labels.
% Specially, each $x_i$ is paired with $\widetilde{y_i}$, which are randomly sampled from label space ${\{y_i\}_{i=1}^k}$.

% \paratitle{Task Learning.}
$\bullet$~\emph{Abstract}:
To evaluate TL ability, we map the original labels in demonstrations to semantically unrelated ones (\eg numbers, letters, or symbols).

With the above settings, we can conduct the corresponding empirical experiments by manipulating the input-label relations (\ie \emph{random} and \emph{abstract} settings) to quantify the effect of TL and TR. 
% More experimental details about the above three settings can be found in Appendix~\ref{}. 
% , which can eliminate the effect of TR.
% This evaluation method is called the \textit{abstract} setting.

\ignore{On the other hand, task learning (TL) refers to learning a new mapping from input-label pairs through demonstrations.
The main difference between the dual abilities is that TL requires correct input-label pairs, allowing the model to learn the correct mapping from examples.
Following \citet{Disentangle-ACL-2023}, when evaluating TL ability, we construct 1-1 mapping to randomly map each label in the label space to a specific semantically unrelated and abstract label (\eg numbers, letters, and symbols).
Specifically, we map each label $y_i$ in the examples to $\phi(y_i)$ using the mapping $\phi: \mathcal{Y} \to \mathcal{Y^*}$, and predict $\phi(\hat{y}_{k+1})$ as the target label.
In this paper, we use abstract symbols to reflect TL ability.
More types of abstract labels are detailed in Appendix~\ref{app-sub:more-abstract}.}

% On the other hand, 任务学习指的是利用上下文示例中的input-label pair进行学习，建模它们之间新的映射关系。
% 它与任务识别之间最大的不同是需要正确的input-label pair，以此让模型从示例中学习正确的映射关系。
% Following ~\citet{Chen}, 在评测任务学习能力时，我们将label space中的label与其他语义无关 and abstract的标签进行随机1-1映射 (\eg numbers, letters and symbols).
% Specially, 我们通过映射 $\phi: \mathcal{Y} \to \mathcal{Y^*}$ 将所有的label $y_i$ 映射成 $\phi{y_i}$，并将 $\phi{\hat{y}_{k+1}}$ 作为target label进行预测.
% In this paper, we use abstract symbols to reflect TL ability.
% More types of Abstract Labels are detailed in ~\ref{}.

\subsection{Competition Measurement}
\ignore{We adopt a sample-based evaluation protocol.
For each test sample, we randomly sampled \textcolor{red}{5} demonstrations from the training set and calculated their performance on different tasks and the average accuracy as the result for this data point.}

In this paper, an important hypothesis is that
competitive relationships exist between TR and TL during pre-training. 
To investigate this, we assume that the intermediate checkpoints of LLMs are available, denoted as $\mathcal{M}_{\theta} = \{{M}_{\theta_1}, {M}_{\theta_2}, \cdots, {M}_{\theta_t}\}$.  
%we propose several quantitative metrics to observe the intermediate checkpoints of LLMs $\mathcal{M}_{\theta} = \{{M}_{\theta_1}, {M}_{\theta_2}, \cdots, {M}_{\theta_t}\}$.
Specifically, we first calculate the performance change of TR and TL during pre-training as follows:
\begin{align}
    \Delta \text{TR}_i &= \text{Acc}_{i+1}^{\text{rand}} - \text{Acc}_{i}^{\text{rand}}, \\
    \Delta \text{TL}_i &= \text{Acc}_{i+1}^{\text{abs}} - \text{Acc}_{i}^{\text{abs}},
\end{align}
% where $\text{Acc}_{i}^{\text{rand}}$ and $\text{Acc}_{i}^{\text{abs}}$ denote the accuracy under the random and abstract settings at the $i$-th step.
where $\text{Acc}_{i}^{\text{rand}}$ and $\text{Acc}_{i}^{\text{abs}}$ denote the accuracy of the intermediate checkpoint ${M}_{\theta_i}$ under the \emph{random} and \emph{abstract} settings introduced in Section~\ref{sec-background}.

If the performance of TR and TL changes in  opposite directions, it would indicate that competition actually occurs, which can be represented as:
\begin{align}
    C_i^h ={}& \mathbb{I}(\Delta \text{TR}_i \cdot \Delta \text{TL}_i < 0) \notag \\
             & \cdot \mathbb{I}(\left| \Delta \text{TR}_i \right| > \epsilon) \cdot \mathbb{I}(\left| \Delta \text{TL}_i \right| > \epsilon), \label{eq:competition-indicator}
\end{align}
where $\mathbb{I}(\cdot)$ is the indicator function.
Here, we consider two additional indicator functions to reduce the influence of inaccurate performance estimation.
$\epsilon$ is set to {0.01} in our experiment.
Furthermore, to measure the intensity of competition $C_i^s$, we consider using the ratio between the performance changes of TR and TL:
\begin{align}
    C_i^s = C_i^h \cdot & \left[ \mathbb{I}(\Delta\text{TR}_i < 0) \cdot \left| \frac{\Delta\text{TR}_i}{\Delta\text{TL}_i} \right| \right. \notag \\
             & \left. +\ \mathbb{I}(\Delta\text{TL}_i < 0) \cdot \left| \frac{\Delta\text{TL}_i}{\Delta\text{TR}_i} \right| \ \right].
\end{align}
Here, we assume that an increase in the performance of one ability at the cost of a decrease in the performance of the other ability indicates the intensity of competition.
A larger value of $C_i^s$ suggests more intense competition, as it implies a greater decrease in the performance of one ability for a given increase in the performance of the other.
Moreover, to investigate the dynamics of competition during pre-training, we calculate the cumulative intensity score $R_i$ as follows:
\begin{equation}
    R_i = \frac{\sum_{j=1}^i {C_j^s}}{\sum_{j=1}^N {C_j^s}}.
\label{eq:competition-evolving}
\end{equation}
This measure tracks the cumulative proportion of the total competition intensity up to the $i$-th training step, providing insight into how competition evolves over time.

\ignore{\begin{equation}
    I_i = 
    \begin{cases} 
    1 & \text{if } \Delta \text{TR}_i \cdot \Delta \text{TL}_i < 0 \\
    0 & \text{otherwise}
    \end{cases}
\end{equation}

However, $I_i$ only considers the presence of competition.
It does not measure the intensity of competition. 
Therefore, we use the absolute value of the performance changes of the two abilities $C_i$ to reflect the intensity of competition at the $i$th step.
% 然而，$I_i$ 的定义仅考虑了竞争出现与否，并不能很好的度量竞争的剧烈程度。因此，我们将两种能力性能变化的绝对值 $C_i$ 来反映两种能力在第$i$个checkpoint竞争的剧烈程度。

\begin{equation}
    C_i = 
    \begin{cases} 
    |\Delta \text{TR}_i - \Delta \text{TL}_i| & \text{if } I_i = 0 \\
    0 & \text{if } I_i = 1 
    \end{cases}
\end{equation}

Furthermore, to measure the competition during the pre-training process, we introduce hard and soft competitiveness.
They are defined as the average occurrence and performance of competition.
% 因此，我们定义hard competitiveness为训练过程中竞争出现次数的平均值。
% 相似的，我们定义soft competitiveness为训练过程中竞争性能的平均值。

\begin{equation}
\begin{split}
    \text{Hard\_com} = \frac{1}{N} \sum_{i=1}^N I_i \\
    \text{Soft\_com} = \frac{1}{N} \sum_{i=1}^N C_i
\end{split}
\end{equation}

To further consider the competition intensity changes during the pre-training process, we quantitatively calculate the percentage of all competition quantities before the current step and introduce $CP_i$.
% 为了进一步考虑竞争剧烈程度随预训练变化的情况，我们定量计算了当前步数之前的所有竞争量占总竞争量的百分比并引入了$CP_i$

\begin{equation}
    CP_i = \frac{\sum_{k=1}^i {C_k}}{\sum_{k=1}^N {C_k}}
\end{equation}}

\section{Empirical Analysis}
In this section, we present the empirical analysis of the competition relationships between TR and TL for ICL.

\subsection{Experimental Setup}
\label{subsec:exp}

\paratitle{Tasks and Datasets.}
Following \citet{Disentangle-ACL-2023}, we select 16 datasets across four types of tasks for the experiment: sentiment analysis, topic/state classification, toxicity detection, and natural language inference/paraphrase detection.
Details about the datasets are depicted in Appendix~\ref{app:detailed-exp-settings}.
Due to computational constraints, we sample 1000 examples from each dataset for evaluation.

\paratitle{Models.}
Since our work focuses on the pre-training dynamics of ICL, we select LLMs that have more than 350M parameters and provide their intermediate checkpoints: the Pythia suite~(6 model sizes ranging from 410M to 12B)~\cite{Pythia-model}, MiniCPM-2B~\cite{Minicpm-model}, Baichuan2-7B~\cite{Baichuan2-model}, Amber-7B~\cite{LLM360-model}, CrystalCoder-7B~\cite{LLM360-model}, and OLMo-7B~\cite{Olmo-model}.
Due to computational constraints, we sample 16 checkpoints in addition to the final one.
They are evenly distributed in the pre-training process.
Experiments with other numbers of checkpoints yield similar results, which are shown in Appendix~\ref{app-sub:more-ckpt}.
To make the output as deterministic as possible, we set \texttt{temperature=0}.

\paratitle{Other Details.}
% We uniformly sample 17 checkpoints (\ie 16 competing quantities) during pre-training for in-depth analysis. 
% More discussions on the number of checkpoints are provided in Appendix~\ref{app-sub:more-ckpt}.
We use 16 randomly sampled examples as demonstrations by default across the paper following \citet{Rethinking-2022-EMNLP}.
The discussion about the number of examples can be found in Appendix~\ref{app-sub:more-demo}.
We use minimal templates to construct demonstrations following \citet{Disentangle-ACL-2023}.
Specifically, we use a single newline character (\ie \textbackslash n) to connect each input-label pair and three ones to separate examples.
We utilize symbols as labels in the abstract setting. Other kinds of abstract labels yield similar results as discussed in Appendix~\ref{app-sub:more-abstract}.
The results are averaged across five random seeds.
% We provide more detailed example templates in Appendix~\ref{app:detailed-exp-settings}.

% 我们根据预训练过程中两个checkpoint间隔的token数，均匀采样了预训练中间过程中17个checkpoint(\ie 16个竞争量)进行深入分析，更多关于预训练中间过程中checkpoint数量的讨论我们放在了Appendix~\ref{}。
% 我们默认使用16-shot作为我们的实验设置，more关于示例数量的讨论请参见 Appendix~\ref{}.
% Following ~\citet{rethinking}, 我们使用minimal templates to form the sequences from demonstrations.
% Specially, 我们使用"\n"连接每一个input-label pair, 并使用"\n\n\n"作为示例之间的连接符。
% 我们把更详细的示例模版放在 Appendix~\ref{}.
\subsection{Task Recognition and Task Learning Are Competitive During Pre-Training}
\label{sec-main_res}

\figurename~\ref{fig:performance-three-setting} shows that the emergence of ICL meets many fluctuations, along with competition between its two abilities (\ie TR and TL).
In this section, we delve into this competition and unveil its relationship with ICL.

\begin{figure}[t]
    \centering
    \includegraphics[width=\columnwidth]{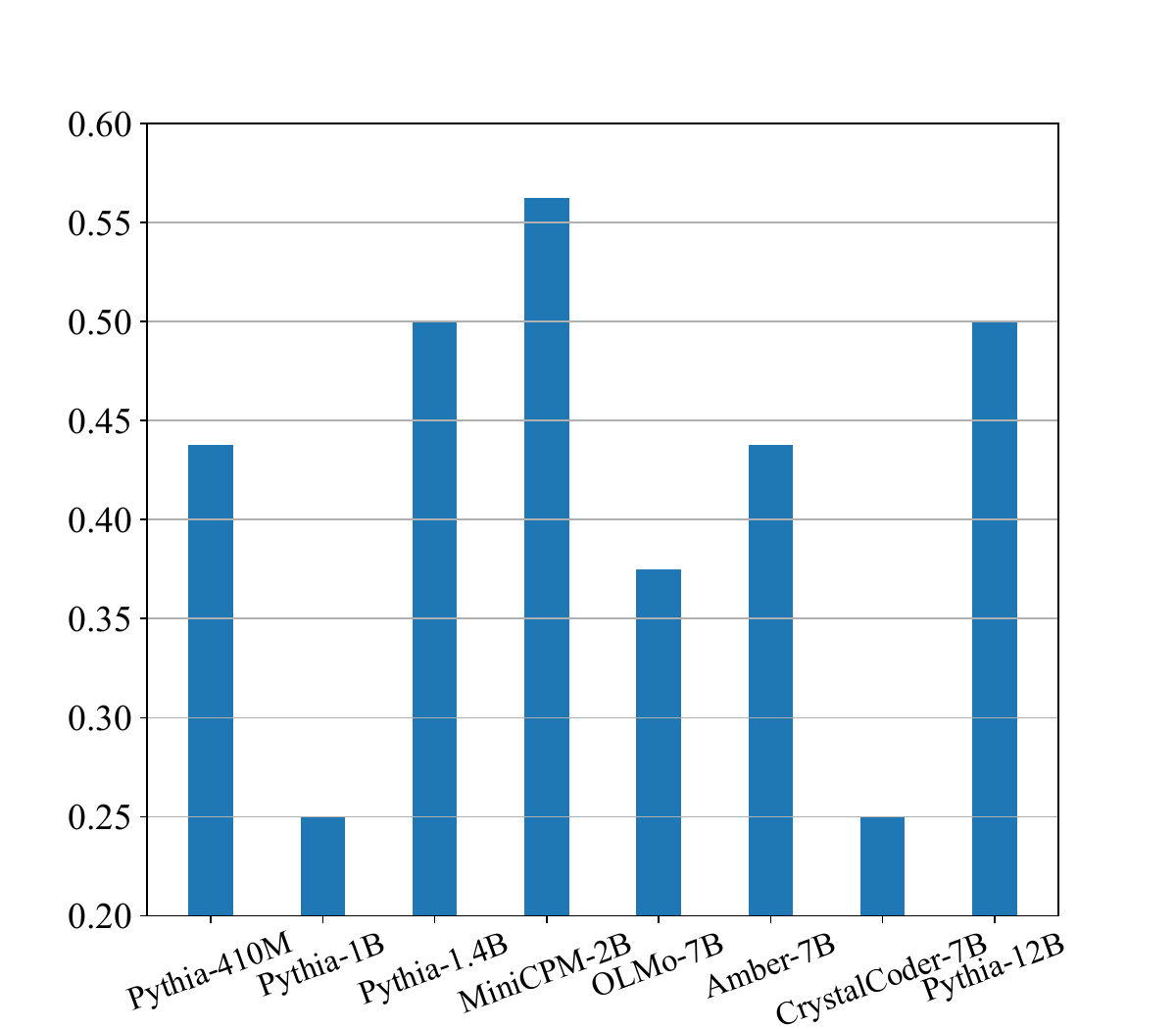}
    \caption{Average ratio of competition for LLMs.}
\label{fig:existence}
\end{figure}

\paratitle{The Existence of Competition.}
To confirm the existence of competition between TR and TL, we investigate the pre-training process of 8 LLMs with various training settings.
Specifically, we calculate the average ratio of competitions according to the indicator metric $C_i^h$ defined in Eq.~\eqref{eq:competition-indicator}.
As illustrated in \figurename~\ref{fig:existence}, all the LLMs exhibit certain levels of competition during pre-training.
For some LLMs, even more than half the time there exists competition.
It suggests that the competition between TR and TL is a widespread phenomenon during pre-training.

\ignore{To more clearly present our findings, we calculated the hard competitiveness metric $C^h_i$ across \textcolor{red}{X} models.
As illustrated in Figure~\ref{fig:existence}, the dual abilities of all models exhibit competition during the pre-training process.
Specifically, Pythia-2.8B has competition in over half of the entire pre-training process.}

% 为了更清晰的展现我们的发现，我们计算了在X个模型的hard competitiveness的指标。
% As illustrated in Figure~\ref{fig:existence}, 所有的模型的dual abilities在预训练过程中都会出现竞争的现象。
% 特别地，XXX模型在预训练过程中发生竞争的占比甚至超过了全部预训练过程的一半以上。

% 算golden的性能和竞争关系的相关程度（和参数量相关性，数据量相关性对比）

% \paratitle{The Dynamic of Competition.}
% \begin{figure}[t]
%     \centering
%     \begin{subfigure}[b]{0.49\linewidth}
%         \centering
%         \includegraphics[width=\textwidth]{figures/golden-competitiveness-minicpm.pdf}
%         \caption{MiniCPM-2B}
%         \label{fig:minicpm-cis}
%     \end{subfigure}
%     \begin{subfigure}[b]{0.49\linewidth}
%         \centering
%         \includegraphics[width=\textwidth]{figures/golden-competitiveness-olmo.pdf}
%         \caption{OLMo-7B}
%         \label{fig:olmo-cis}
%     \end{subfigure}
%     \caption{The gold accuracy and cumulative intensity score during pre-training process.}
% \label{fig:dynamic}
% \end{figure}

\begin{figure}[t]
    \centering
    \begin{subfigure}[b]{0.49\linewidth}
        \centering
        \includegraphics[width=\linewidth]{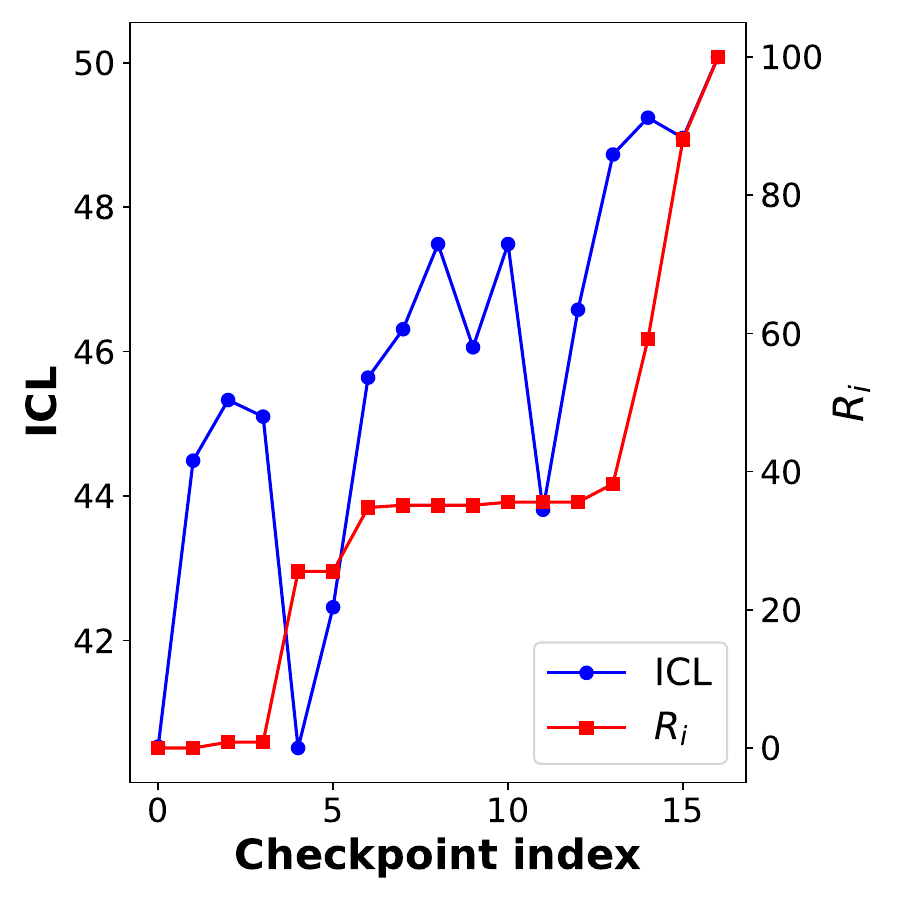}
        \caption{MiniCPM-2B}
        \label{fig:minicpm-cis}
    \end{subfigure}
    % \begin{subfigure}[b]{\linewidth}
    %     \centering
    %     \includegraphics[width=\textwidth]{figures/golden-competitiveness-olmo.pdf}
    %     \caption{OLMo-7B}
    %     \label{fig:olmo-cis}
    % \end{subfigure}
    \begin{subfigure}[b]{0.49\linewidth}
        \centering
        \includegraphics[width=\linewidth]{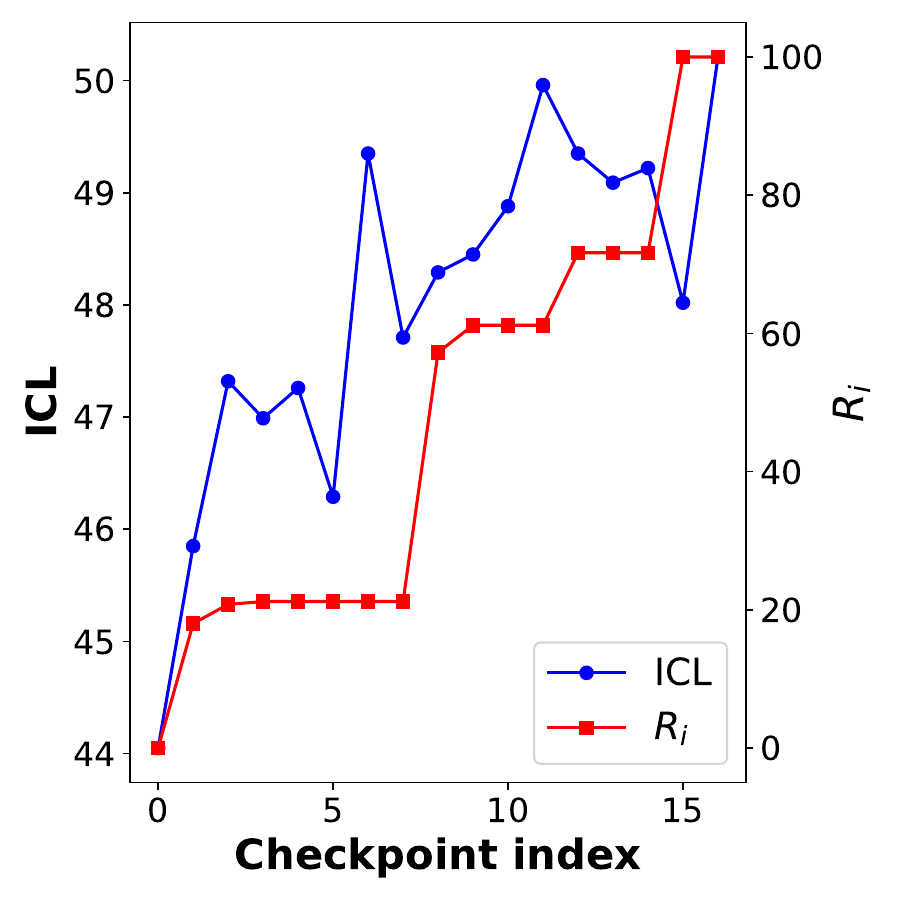}
        \caption{Amber-7B}
        \label{fig:amber-cis}
    \end{subfigure}
    \caption{The performance of ICL and the evolution of competition ($R_i$) during the pre-training of MiniCPM-2B and Amber-7B.}
\label{fig:dynamic}
\end{figure}

% \begin{figure}[t]
%     \centering
%     \includegraphics[width=\columnwidth]{figures/golden-competitiveness-minicpm.pdf}
%     \caption{The performance of ICL and the evolution of competition ($R_i$) during the pre-training of MiniCPM-2B.}
% \label{fig:dynamic-minicpm}
% \end{figure}

\paratitle{The Dynamic of Competition.}
We further explore the intensity of competition and its evolution during pre-training.
Specifically, we choose MiniCPM-2B and Amber-7B, which are trained with over a trillion tokens with different amounts of parameters.
We track the evolution of the intensity of competition using the metric $R_i$ defined in Eq.~\eqref{eq:competition-evolving}.
Results are shown in \figurename~\ref{fig:dynamic}.
We can observe that the intensity of competition typically repeats the ``stable--rise'' pattern, which usually corresponds to the fluctuation and increase in the performance of ICL.
Such an interesting phenomenon inspires us to further examine the relationship between the competition and the performance of ICL.

% In this section, 我们探究与训练过程中竞争量的变化情况，及其对ICL性能的影响。
\ignore{\paratitle{Changes of competition during pre-training.}
To delve into the dynamics of competition during pre-training, we conduct an in-depth analysis of the changes in competition during the pre-training process.
Figure~\ref{} shows the changes in the cumulative intensity scores of Minicpm-2B and OLMo-7B.
We observe that competition mainly follows a two-stage upward trend.
Specifically, both abilities initially increase steadily, then begin to compete and reallocate resources. 
Then, it is followed by a similar process and eventually stabilizes.}

% 为了深入探究预训练过程中竞争的动态，我们对预训练过程中竞争量的变化情况进行深入分析。
% In this part, 我们选择minicpm和olmo两个模型进行进一步分析。
% Figure~\ref{} 展现了这两个模型cumulative intensity score的变化情况。
% 我们可以发现竞争主要有一种两阶段上升的趋势。
% Specially, 首先这两种能力平稳上升，而后两种能力开始竞争而重新分配资源。接着，有进行了一次相似的过程，最终趋于平稳。

\paratitle{The Relationship Between Competition and ICL.}
% \paratitle{The Impact of Competition on ICL Performance}
% Pearson相关系数: -0.61，p值：0.03
\begin{figure}[t]
    \centering
    \includegraphics[width=\linewidth]{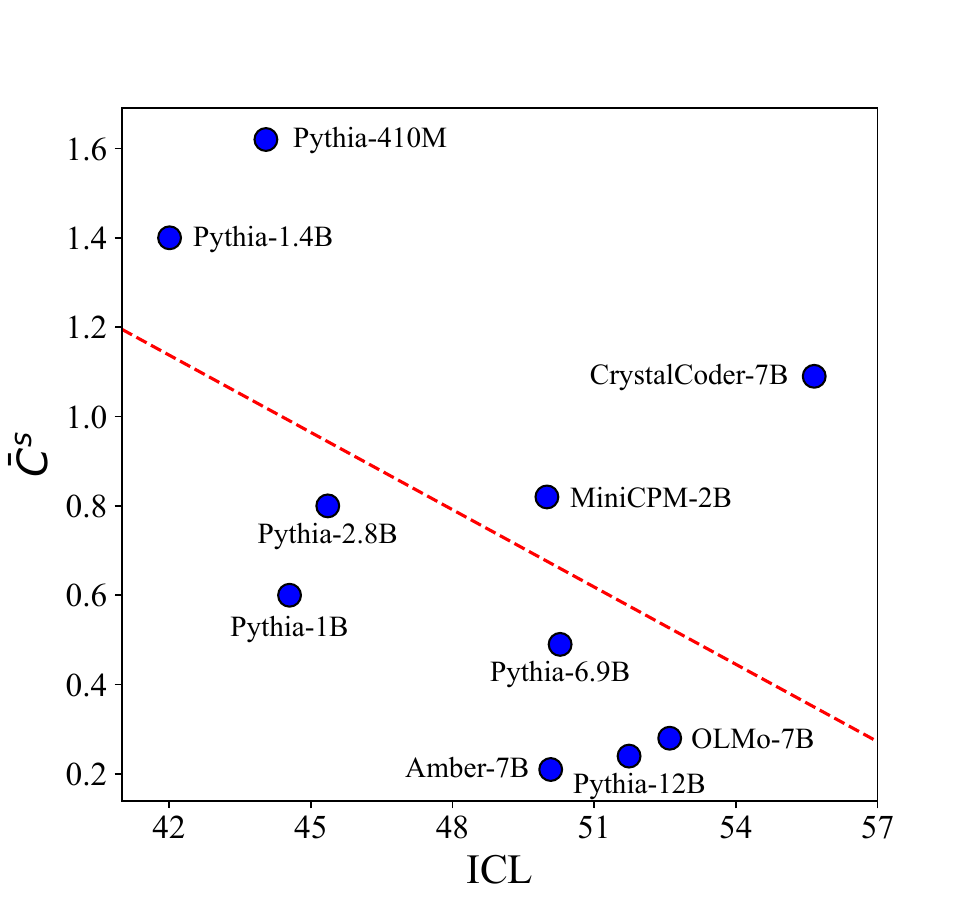}
    \caption{ICL performance of the final checkpoint and the average intensity of competition ($\bar{C}^s$) for LLMs.}
\label{fig:relationship}
\end{figure}
We first examine the relationship between the competition and the performance of ICL based on pre-training dynamics.
As shown in \figurename~\ref{fig:dynamic}, when there exists competition, the performance of ICL tends to increase (78\% of the time for MiniCPM-2B and 57\% for Amber-7B).
However, when there is no competition, the performance of ICL shows fluctuations, making the situation complicated.
To make it clear, we shift our perspective to the entire pre-training process.
Specifically, we examine the relationship between the average intensity of 
competition $\bar{C}^s$ defined in Eq.~\eqref{eq:competition-evolving} and the ICL performance of the final checkpoint.
As illustrated in \figurename~\ref{fig:relationship}, with the increase of $\bar{C}^s$, the ICL performance tends to drop, with the exception of MiniCPM-2B and CrystalCoder-7B (they will be discussed in Section~\ref{sec-detailed_analysis}).
To further verify their correlation, we calculate the Pearson correlation coefficient.
The result is {-0.591}, validating their negative correlation.
% In summary, our analysis demonstrates a clear negative correlation between the performance of ICL and the competition between TR and TL.
This finding has important implications for optimizing pre-training processes and suggests that managing the competition between TR and TL could be crucial for enhancing ICL ability.

\ignore{After discovering the phenomenon of competition changing during the pre-training process, we aim to explore the relationship between the ICL ability of the current step $\text{Acc}^{\text{golden}}_i$ and the intensity of competition $C_i^s$.
As illustrated in Figure~\ref{}, we find that...
Finally, we examine the correlation between the changes in performance under the golden setting and the intensity of competition for each checkpoint. We find that the correlation coefficient reaches $X$, indicating a strong negative correlation between the two.}

% After 发现了竞争在预训练过程中变化的现象，我们旨在探究当前step的ICL能力$\text{Acc}^{\text{golden}}$和竞争强度intensity of competition$C_i^s$之间的关系。
% 
% As illustrated in Figure~\ref{}, 我们发现

% 最后，我们examine 每一个checkpoint之间 the correlation between the golden setting下性能的变化在和 intensity of competition 之间的关系。
% 我们发现相关系数能达到$X$，这表明了这二者之间存在极强的负相关性。
\subsection{How Do Factors of Pre-Training Influence the Competition?}
\label{sec-detailed_analysis}

As discussed in Section~\ref{sec-main_res}, the competition between TR and TL during pre-training demonstrates a strong correlation with the final ICL performance. 
This motivates us to investigate the influence of pre-training factors on the competition level. 
Specifically, we investigate several common factors, \ie model size, dataset size, and data curriculum.
\ignore{After discovering that the dual abilities of ICL exhibit a significant competitive relationship during the pre-training process, which affects the final ICL performance.
In this section, we delve into how pre-training factors (\ie model size, dataset size, and data curriculum) influence the competition.
}

% 在发现了dual abilities of ICL在预训练过程中存在明显的竞争关系，从而影响最终的ICL performance之后。
% 在这一章节，我们深入考虑了预训练的因素（\ie 模型规模，数据规模，数据课程）是如何影响竞争的。

% \input{tables/model-size}

\subsubsection{Effect of Model Size}

\begin{figure}[t]
    \centering
    \begin{subfigure}[b]{0.49\linewidth}
        \centering
        \includegraphics[width=\linewidth]{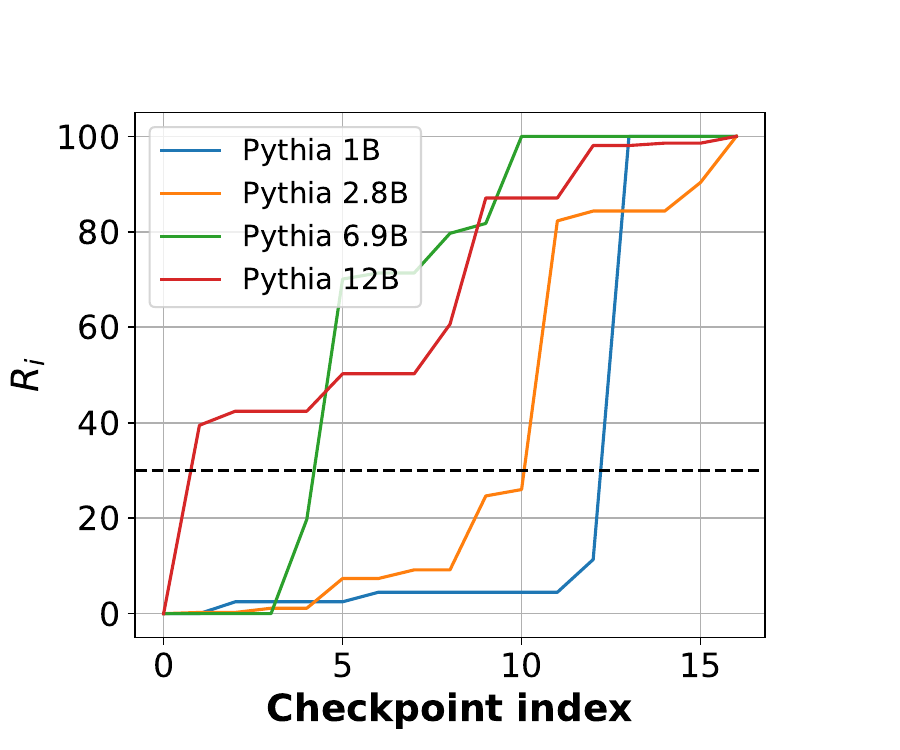}
        \caption{Evolving process}
        \label{fig:model-size-r}
    \end{subfigure}
    \begin{subfigure}[b]{0.49\linewidth}
        \centering
        \includegraphics[width=\linewidth]{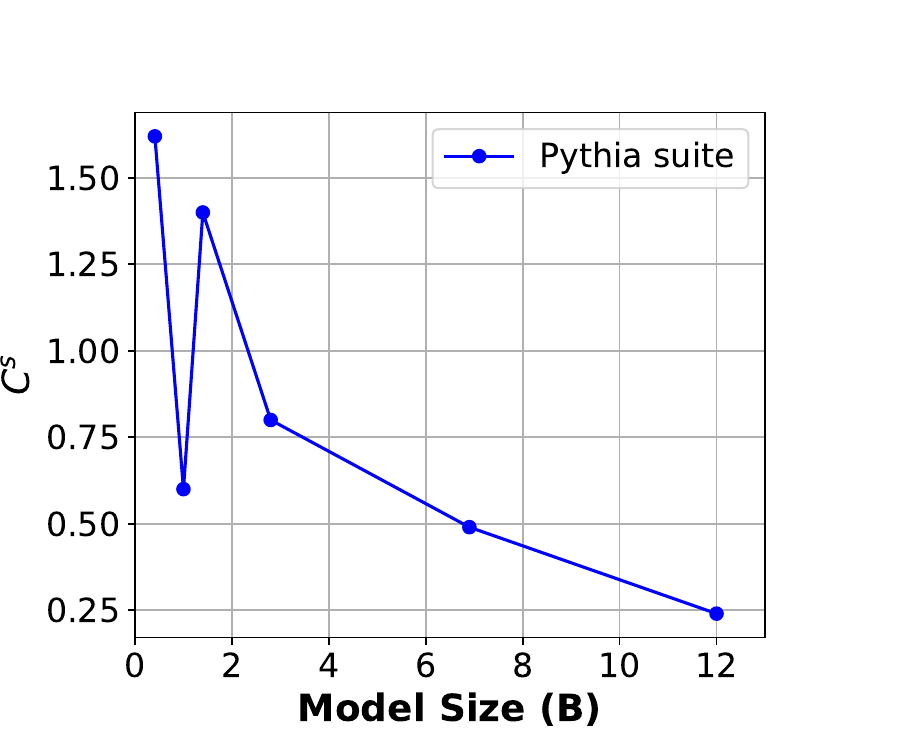}
        \caption{Average intensity}
        \label{fig:model-size-soft}
    \end{subfigure}
    \caption{The evolution and average intensity ($\bar{C_i^s}$) of competition of LLMs with different model sizes.}
\label{fig:model-size}
\end{figure}

% \begin{figure}[t]
%     \centering
%     \includegraphics[width=\linewidth]{figures/model-size-avg-soft.pdf}
%     \caption{The effect of model size.}
% \label{fig:model-size}
% \end{figure}

We investigate the effect of model size on the competition between TR and TL.
Specifically, we use the Pythia suite
% \footnote{We exclude Pythia-70M and Pythia-160M because their ICL performance is worse than the random guess.} 
for experimentation since these models share the same training setting in addition to the number of parameters.

We first pay attention to the differences in the evolution of competition.
We can observe from Figure~\ref{fig:model-size-r} that as the model size increases, the evolving curve of competition keeps moving to the left.
This suggests that scaling up model size could make the appearance of competition appear earlier.
One possible reason is that the learning ability of larger LLMs is stronger, and they can possess TR and TL more quickly, thus causing competition between them to occur earlier.

Then, we focus on the changes in the average competition intensity.
As illustrated in Figure~\ref{fig:model-size-soft}, the average competitive intensity sharply decreases with the increase of model size, with the exception of Pythia-1B.
This indicates that scaling up the model size is helpful in reducing the overall competition.
This may be attributed to the fact that LLMs with more parameters have a larger capacity, where TR and TL can be allocated with more exclusive resources (\eg neurons).
As a result, although the competition becomes earlier in larger LLMs, the average intensity of competition becomes lower.
Interestingly, we can observe that overall, the average intensity of competition scales as a power-law with model size, which follows a similar pattern with the training loss in the scaling law of LLMs~\cite{kaplan2020scaling}.
We leave the exploration of the relationship between the scaling law of LLMs and the competition between TR and TL as future work.

% 为了深入探究模型规模对于上下文学习竞争量的影响，在这一部分，我们使用Pythia系列的6个模型ranging from 410M to 12B，它们拥有相同的模型结果以及相同的训练数据和一致的训练顺序.~\footnote{我们排除Pythia系列中两个最小的模型(\ie Pythia-70M and Pythia-160M), 因为他们ICL的性能不如random guess。}
% Figure~\ref{fig:model-size}中呈现了模型大小和average intensity of competition的关系。

% It can be observed that模型的规模越大，平均竞争量越少，这表明了模型规模的扩大可以缓解预训练过程中两种能力的竞争。
% 一种可能的原因是，较大的模型拥有更多的参数，这使得它们能够更好地捕捉和表示复杂的特征。这些额外的参数允许模型在处理任务时具有更大的灵活性和表达能力，从而可以更好地平衡任务识别和任务学习之间的需求。通过持续的预训练，模型内部神经元分化更为明显，因此这两种能力可以在更大的模型中更有效地共享资源而不相互干扰，从而减少任务识别和任务学习之间的竞争。

% 竞争强度，方程

% pythia, map-neo

\subsubsection{Scaling Dataset Size}

% \begin{figure}[t]
%     \centering
%     \begin{subfigure}[b]{0.49\linewidth}
%         \centering
%         \includegraphics[width=\textwidth]{figures/small_model_token.pdf}
%         \caption{Pythia-2.8B and MiniCPM-2B}
%         \label{fig:minicpm-preformance}
%     \end{subfigure}
%     \begin{subfigure}[b]{0.49\linewidth}
%         \centering
%         \includegraphics[width=\textwidth]{figures/large_model_token.pdf}
%         \caption{Pythia-6.9B, Amber-7B, OLMo-7B}
%         \label{fig:olmo-preformance}
%     \end{subfigure}
%     \caption{The impact on dataset size.}
% \label{fig:dataset-size}
% \end{figure}

\begin{figure}[t]
    \centering
    \includegraphics[width=0.49\linewidth]{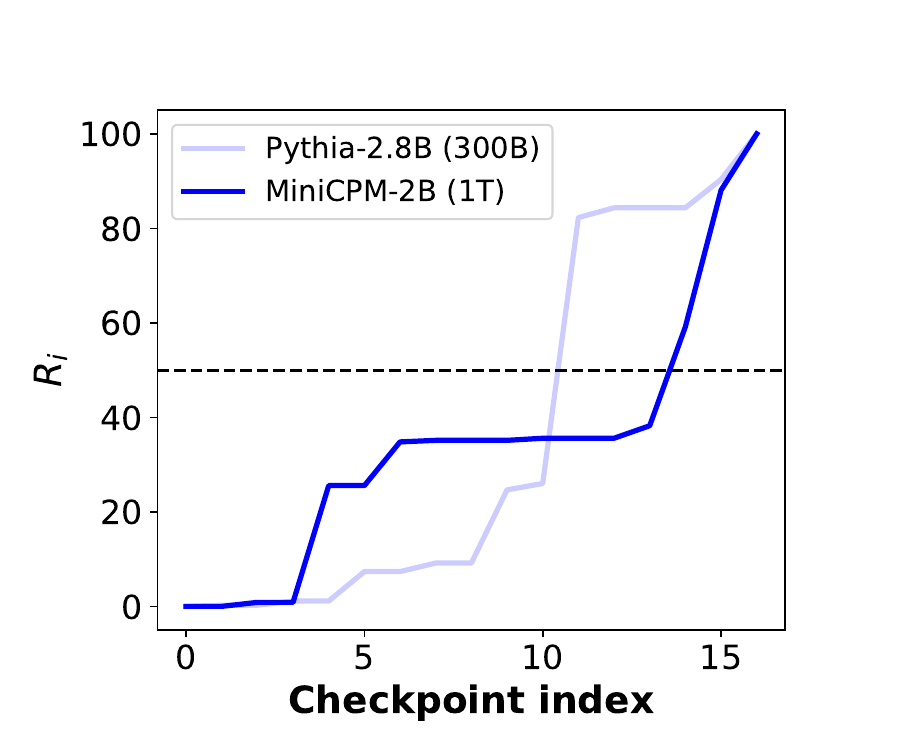}
    \includegraphics[width=0.49\linewidth]{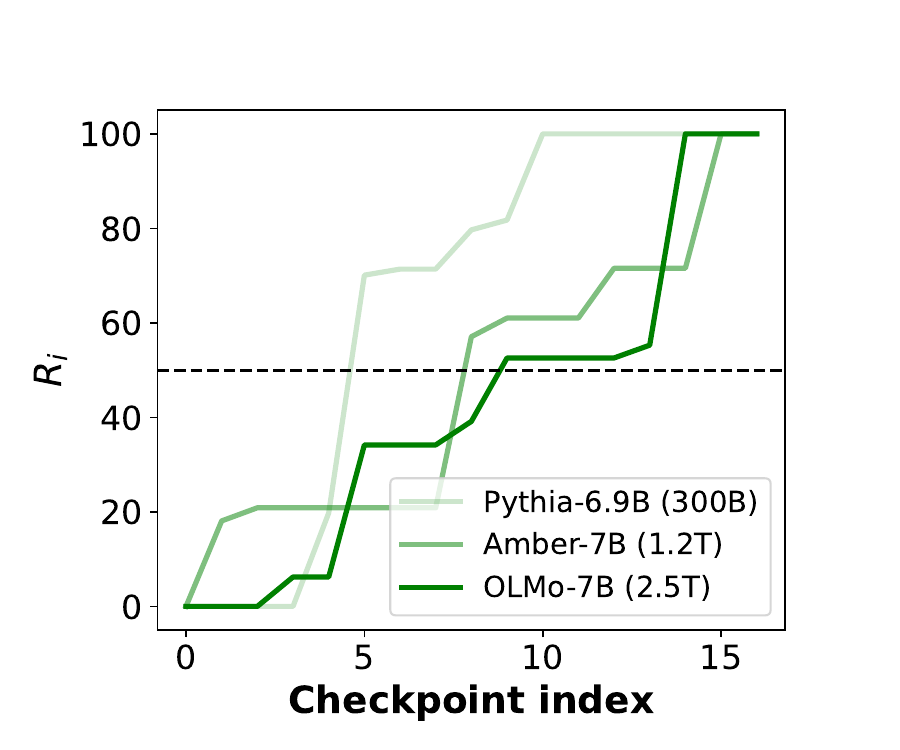}
    \caption{Competition evolving process ($R_i$) of LLMs trained with different dataset sizes.}
    \label{fig:dataset-size}
\end{figure}

% \begin{figure}[t]
%     \centering
%     \includegraphics[width=\linewidth]{figures/large_model_token.pdf}
%     \caption{
%         Competition evolving process ($R_i$) of LLMs trained with different dataset sizes.
%     }
%     \label{fig:dataset-size-large}
% \end{figure}

In this part, we explore the impact of dataset size on the competition between TR and TL.
We conduct experiments using models with roughly the same number of parameters but trained with different dataset sizes.
Specifically, we make the comparison among two sets of LLMs: (Pythia-2.8B and MiniCPM-2B) and (Pythia-6.9B, Amber-7B, and OLMo-7B).

\figurename~\ref{fig:dataset-size} illustrates the evolution of competition during pre-training for these two sets of LLMs.
It can be observed that, for both sets of LLMs, the evolving curve keeps moving to the right with the increasing of dataset size.
This suggests that scaling up dataset size could postpone the competition.
The possible reason behind this is that, when pre-trained on a small dataset, LLMs can quickly memorize the knowledge contained in the dataset.
Thus, they can develop the TR ability for performing ICL in an early stage.
Meanwhile, the TL ability can also be easily acquired, as it primarily involves directly utilizing the information in context, as discussed by \citet{singh2024transient}.
As a result, the competition between TR and TL occurs in the early stage of pre-training.
With the increase in dataset size, the development of the TR ability becomes slower since there is more knowledge required to memorize.
Therefore, more competition happens at a later time, which makes the evolving curve shift to the right.

% In this part, 我们探究dataset size对于竞争的影响。
% 我们开展了两种相同规模 (\ie 2.8B and 7B)下，不同数据量的模型进行实验。
% Figure~ref{dataset-size} 说明了我们的实验结果。
% We find that 随着预训练数据量的增加，总体竞争的趋势出现滞后的趋势。换句话说，预训练数据量越大，竞争出现在预训练更后面的阶段。
% 当数据量较小时，模型在预训练的早期阶段就会迅速地从有限的数据中提取出大量信息并开始对任务进行区分。这种快速的学习导致模型在预训练初期就出现了显著的竞争关系。
% 然而，随着数据量的增加，模型在预训练初期需要处理和学习的数据量大幅增加，这使得模型在初期阶段主要集中在基础知识的积累和初步的任务识别。
% 在预训练的后期，当模型已经积累了足够的知识并开始精细化其内部表示时，竞争才逐渐显现。

% pythia-1b, neo, minicpm
% pythia-6.9b, amber, olmo

% \subsection{Adjusting Data Mixture}

% amber, pythia-6.9b, olmo

\begin{figure}[t]
    \centering
    \includegraphics[width=\linewidth]{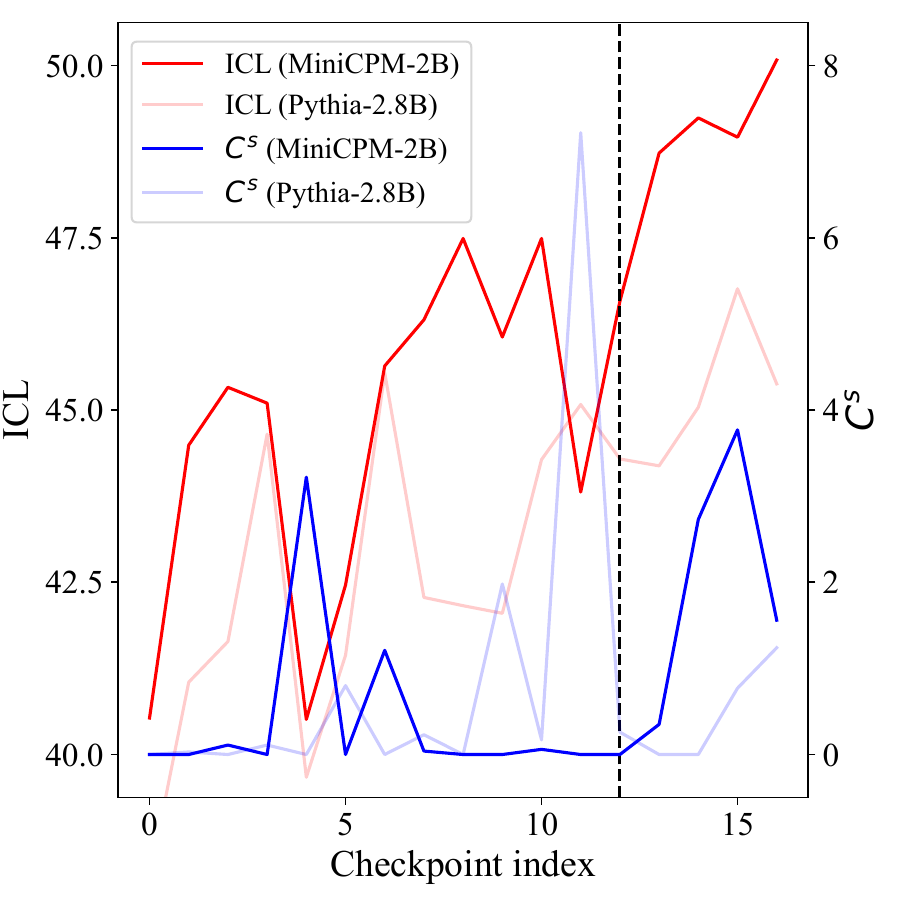}
    \caption{ICL performance and competition intensity ($C^s$) of MiniCPM-2B and Pythia-2.8B. The dashed line is used to distinguish between different training stages.}
\label{fig:minicpm-curriculum}
\end{figure}
\begin{figure}[t]
    \centering
    \includegraphics[width=\linewidth]{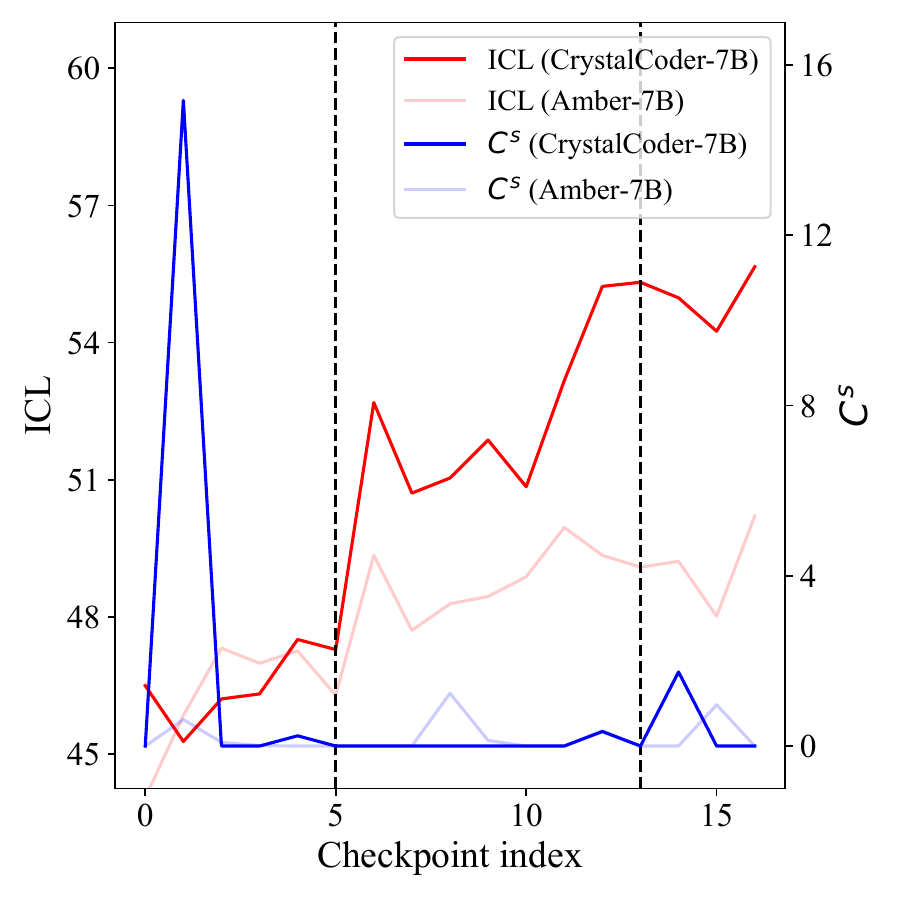}
    \caption{ICL performance and competition intensity ($C^s$) of CrystalCoder-7B and Amber-7B. Dash lines are used to distinguish between different training stages.}
\label{fig:coder-curriculum}
\end{figure}

\subsubsection{Scheduling Data Curriculum}

In this part, we explore the influence of data curriculum on the competition between TR and TL.
Here, we consider two representative strategies for scheduling data curriculum: (1) quality curriculum, which makes arrangements for data of different qualities, and (2) domain curriculum, which makes arrangements for data from different domains.

We first pay attention to the influence of quality curriculum on the competition.
Specifically, we use MiniCPM-2B for experimentation, which utilizes coarse-quality unlabeled data in the first stage and mixes high-quality labeled data in the second stage.
We compare its pre-training process with that of Pythia-2.8B, which has a similar model size and dataset size.
As illustrated in Figure~\ref{fig:minicpm-curriculum}, the ICL performance of MiniCPM-2B traps in fluctuations in the latter half of the first stage while starting to increase again in the second stage.
Meanwhile, the competition in MiniCPM-2B also becomes active in the second stage, and its intensity rapidly increases.
In contrast, the ICL performance of Pythia-2.8B keeps in fluctuation in the later stages of pre-training, and the competition is relatively less active.
% It suggests that quality curricula can increase the intensity of competition to enhance the performance of ICL.
One possible reason for the success of quality curriculum is that part of the knowledge in high-quality labeled data is actually covered by large-scale coarse-quality data, which may cause competition between TR and TL to enhance their learning for the high-quality data.

We then focus on the influence of domain curriculum on the competition.
Specifically, we use CrystalCoder-7B for experimentation, which utilizes general domain data (\ie SlimPajama~\cite{cerebras2023slimpajama}) in the first stage, mixes general and code domain data (\ie SlimPajama and StarCoder~\cite{li2023starcoder}) in the second stage, and mainly uses specific programming language data (\ie Python and web-related data sampled from StarCoder) in the final stage.
Similar to the quality curriculum, we compare its pre-training process with that of Amber-7B, which has a similar model size and dataset size.
As shown in Figure~\ref{fig:coder-curriculum}, compared with Amber-7B, CrystalCoder-7B demonstrates much less competition in the second stage, along with a higher performance improvement.
The underlying reason may be the domain difference between general text and code, which brings additional knowledge for LLMs to develop TR and postpone the competition between TR and TL on shared knowledge.
In the final stage of CrystalCoder-7B, the competition becomes active again.
This could be attributed to data duplication since training data is sampled from StarCoder, which has been utilized in the second stage.
Duplicated data can stimulate the competition between TR and TL to strengthen their learning of specific skills or knowledge (\eg specific programming languages for CrystalCoder-7B), thus achieving better model specialization.
\section{From Competition to Collaboration at Inference Time}
\label{sec-method}

%Since the dual abilities of ICL are shown to be competitive during pre-training, as discussed in Section~\ref{sec-main_res}, 
%To mitigate the competition effect between TR and TL abilities (helpful for improving the ICL performance)

As discussed in Section~\ref{sec-main_res},  the competitive relations between TR and TL abilities would lead to a decrease in ICL performance, and our idea is to mitigate the competition effect and facilitate their collaboration at inference time.
%we propose adaptive ensemble learning to mitigate the competition effect and facilitate their collaboration at inference time.
In this section, we first introduce the proposed adaptive ensemble learning method and then show the experimental results.

\ignore{Based on our analysis in Section~\ref{sec-main_res}, we find a strong competitive relationship between the dual abilities of ICL during pre-training.
In this section, we aim to enhance the performance of ICL by fusing both abilities at inference time.}
% 发现ICL两种能力在预训练期间很强的竞争关系。
% 在这一章，我们旨在通过在推理阶段融合上下文学习的两种能力来实现增强ICL的性能。

\subsection{Method}

Our previous analysis in Section~\ref{sec-main_res} shows that although ICL achieves the best performance at the end of training under the gold setting (correct input-label mapping), the corresponding dual abilities usually do not achieve the best performance simultaneously.
This observation suggests that it would achieve better ICL performance if we could integrate the best TR and TL capabilities in one model. 
Based on this idea, we propose to fuse the corresponding model checkpoints with ensemble learning. 
%by fusing different intermediate checkpoints.
%Since different checkpoints are proficient in different sub-abilities, we propose to fuse them with ensemble learning.
Specifically, we first select two checkpoints with the best ability of TR and TL respectively, and then integrate their probability distributions to make the prediction:
\begin{equation}
    \mathop{\arg\max} \limits_{y \in \mathcal{Y}} \left[ w_r \text{Pr}_{{r}}^{\text{rand}}(y | x) + w_l \text{Pr}_{{l}}^{\text{abs}}(y | x) \right],
\end{equation}
where $\text{Pr}_{{r}}^{\text{rand}}(y | x)$ and $\text{Pr}_{{l}}^{\text{abs}}(y | x)$ denote the probability for the TR and TL models to predict the label $y$ under the random and abstract settings respectively, and $w_r$ and $w_l$ denote the weights for the prediction of the TR and TL models respectively.

In addition, considering the contribution of dual abilities to ICL is usually not equal~\cite{}, we further propose an adaptive ensemble learning method for fusion.
Specifically, we control the contribution of each checkpoint by setting the weight according to their performance, which is calculated as follows:
\begin{align}
    w_r &= \frac{\text{Acc}_r^\text{rand} - b}{(\text{Acc}_r^\text{rand} - b) + (\text{Acc}_l^\text{abs} - b)}, \\
    w_l &= \frac{\text{Acc}_l^\text{abs} - b}{(\text{Acc}_r^\text{rand} - b) + (\text{Acc}_l^\text{abs} - b)},
\end{align}
where $\text{Acc}_r^\text{rand}$ is the performance of the model for TR under the random setting, $\text{Acc}_l^\text{abs}$ is the performance of the model for TL under the abstract setting, and $b$ is the performance of random guess.

\subsection{Experimental Setting}

To comprehensively validate the effectiveness of our method, we consider three different combinations of TR and TL models for fusion:
(1) the same model (\ie Pythia-1B),
(2) two models with a similar training setting (\ie Pythia-1B and Pythia-2.8B), and
(3) two models with different training settings (\ie Pythia-1B and MiniCPM-2B).
Each model is selected to play the roles of both TR and TL, respectively.
We select checkpoints with the best performance for the required ability.
Other settings are the same as in Section~\ref{subsec:exp}.

We compare our method with three types of baselines:
(1) the TR or TL model itself used for fusion,
(2) LLMs whose parameters are more than the sum of TR and TL models, and
(3) fusion using the same weight (\ie $w_r=w_l$) (``fixed'' in \tablename~\ref{tab:main-exp}).
All the baselines are tested in the gold setting.

\ignore{We select three combinations of different small language models (\ie the same model (TR: Pythia-1B and TL: Pythia-1B), different models from the same series (TR: Pythia-1B and TL: Pythia-2.8B), and different models (TR: MiniCPM-2B and TL: Pythia-1B)) for fusion.
% 是否需要加上其他结果在附录中？
We choose the models with the strongest TR and TL abilities during pre-training in the development set as our fusion checkpoints.
Other experimental setups follow Section~\ref{subsec:exp}.

We compare \textbf{Adaptive fusion} with three types of baselines:
(1) The golden setting of a single large language model
(2) The random and abstract setting of a single small language model
(3) Fixed fusion method, where the same checkpoint is used for a fair comparison.}

% (1) 规模大的模型的golden setting
% (2) 选取的TR和TL能力最强的小模型在对应evaluation mode的性能
% (3) TR和TL小模型固定融合策略

\begin{table}[t]
    \centering
    \resizebox{\columnwidth}{!}{%
    \begin{tabular}{l|c|c}
        \toprule
        \textbf{Model} & \textbf{\# Params} & \textbf{Acc.} \\
        \midrule
        \midrule
        \multicolumn{3}{c}{\textit{Large models}} \\
        \midrule
        \midrule
        Amber-7B\ICL       & 7B    & 50.08 \\
        Pythia-6.9B\ICL      & 6.9B  & 50.22 \\
        Pythia-12B\ICL      & 12B   & 51.74 \\
        OLMo-7B\ICL        & 7B    & 52.10 \\
        Baichuan2-7B\ICL   & 7B    & 52.77 \\
        CrystalCoder-7B\ICL& 7B    & 55.66 \\
        \midrule
        \midrule
        \multicolumn{3}{c}{\textit{Small models}} \\
        \midrule
        \midrule
        % Pythia-1B\TRGold                            & \multirow{3}{*}{\begin{tabular}[c]{@{}c@{}}1B\end{tabular}}            & 47.47 \\
        % Pythia-1B\TLGold                            &             & 44.63 \\  
        Pythia-1B\ICL                          & 1B        & 44.55 \\
        % \midrule
        % Pythia-2.8B\TRGold                          & \multirow{3}{*}{\begin{tabular}[c]{@{}c@{}}2.8B\end{tabular}}          & 44.28 \\
        % Pythia-2.8B\TLGold                          &           & 42.05 \\
        Pythia-2.8B\ICL                             & 2.8B          & 45.38 \\
        % \midrule
        % MiniCPM-2B\TRGold                      & \multirow{3}{*}{\begin{tabular}[c]{@{}c@{}}2.7B\end{tabular}}        & 45.10 \\
        % MiniCPM-2B\TLGold                      &         & 47.49 \\
        MiniCPM-2B\ICL                         & 2.7B        & 50.08 \\
        \midrule
        \midrule
        \multicolumn{3}{c}{\textit{Fusion of small models}} \\
        \midrule
        \midrule
        % Pythia\TR                           & 1B        & 37.67 \\        
        % Pythia\TL                           & 1B        & 43.50 \\
        % Pythia\TRGold                       & 1B        & 47.47 \\
        % Pythia\TLGold                       & 1B        & 44.63 \\
        % Pythia\ICL                          & 1B        & 44.55 \\
        Pythia-1B\TR + Pythia-1B\TL~(fixed)       & \multirow{2}{*}{\begin{tabular}[c]{@{}c@{}}2B\end{tabular}}    & 56.16 \\
        Pythia-1B\TR + Pythia-1B\TL~(adaptive)    &    & \textbf{56.25} \\
        \midrule
        % Pythia-1B\TR                                & 1B            & 37.67 \\
        % Pythia-1B\TL                                & 1B            & 43.50 \\
        % Pythia-1B\TRGold                            & 1B            & 47.47 \\
        % Pythia-1B\TLGold                            & 1B            & 44.63 \\       
        % Pythia-1B\ICL                               & 1B            & 44.55 \\
        % Pythia-2.8B\TR                              & 2.8B          & 36.06 \\
        % Pythia-2.8B\TL                              & 2.8B          & 45.71 \\
        % Pythia-2.8B\TRGold                          & 2.8B          & 44.28 \\
        % Pythia-2.8B\TLGold                          & 2.8B          & 42.05 \\ 
        % Pythia-2.8B\ICL                             & 2.8B          & 45.38 \\
        Pythia-1B\TR + Pythia-2.8B\TL~(fixed)       & \multirow{4}{*}{\begin{tabular}[c]{@{}c@{}}3.9B\end{tabular}}     & 56.62 \\
        Pythia-1B\TR + Pythia-2.8B\TL~(adaptive)    &      & \textbf{56.83} \\
        Pythia-1B\TL + Pythia-2.8B\TR~(fixed)       &      & 55.23 \\
        Pythia-1B\TL + Pythia-2.8B\TR~(adaptive)    &      & 55.39 \\
        \midrule
        % MiniCPM\TR                          & 2B        & 37.98 \\
        % MiniCPM\TL                          & 2B        & 48.93 \\
        % MiniCPM\TRGold                      & 2B        & 45.10 \\
        % MiniCPM\TLGold                      & 2B        & 47.49 \\
        % MiniCPM\ICL                         & 2B        & 50.08 \\
        % Pythia-1B\TR                        & 1B        & 37.67 \\
        % Pythia-1B\TL                        & 1B        & 43.50 \\
        % Pythia-1B\TRGold                    & 1B        & 47.47 \\
        % Pythia-1B\TLGold                    & 1B        & 44.63 \\
        % Pythia-1B\ICL                       & 1B        & 43.55 \\
        Pythia-1B\TR + MiniCPM-2B\TL~(fixed)      & \multirow{4}{*}{\begin{tabular}[c]{@{}c@{}}3.8B\end{tabular}}   & 55.21 \\
        Pythia-1B\TR + MiniCPM-2B\TL~(adaptive)   &    & 55.31 \\
        Pythia-1B\TL + MiniCPM-2B\TR~(fixed)      &    & 54.31 \\
        Pythia-1B\TL + MiniCPM-2B\TR~(adaptive)   &    & \textbf{55.85} \\
        % \midrule
        % Pythia-2.8B\TR + MiniCPM-2B\TL~(fixed)      & \multirow{4}{*}{\begin{tabular}[c]{@{}c@{}}5.5B\end{tabular}}   & 53.86 \\
        % Pythia-2.8B\TR + MiniCPM-2B\TL~(adaptive)   &    & 54.85 \\
        % Pythia-2.8B\TL + MiniCPM-2B\TR~(fixed)      &    & 54.54 \\
        % Pythia-2.8B\TL + MiniCPM-2B\TR~(adaptive)   &    & \textbf{56.38} \\
        \bottomrule
    \end{tabular}
    }
    \caption{Averaged accuracy across 16 datasets for different models and their fusion. We highlight the highest numbers among fusion with the same model combination.}
    \label{tab:main-exp}
\end{table}

\begin{table}[t]
    \centering
    \small
    \resizebox{\columnwidth}{!}{%
    \begin{tabular}{l|c|c|c}
        \toprule
        \textbf{Models} & \textbf{TR Model} & \textbf{TL Model} & \textbf{Accuracy} \\
        \midrule
        \multirow{4}{*}{\begin{tabular}[l]{@{}l@{}}TR: Pythia-1B \\ TL: Pythia-1B\end{tabular}} 
        & Random & Random & 52.66 \\
        & Best & Random & 53.52 \\
        & Random & Best & 54.19 \\
        & Best & Best & \textbf{56.25} \\
        \midrule
        \multirow{4}{*}{\begin{tabular}[l]{@{}l@{}}TR: Pythia-1B \\ TL: Pythia-2.8B\end{tabular}} 
        & Random & Random & 48.10 \\
        & Best & Random & 50.76 \\
        & Random & Best & 55.42 \\
        & Best & Best & \textbf{56.83} \\
        \midrule
        \multirow{4}{*}{\begin{tabular}[l]{@{}l@{}}TR: Pythia-2.8B \\ TL: Pythia-1B\end{tabular}} 
        & Random & Random & 52.48 \\
        & Best & Random & 53.89 \\
        & Random & Best & 54.39 \\
        & Best & Best & \textbf{55.39} \\
        \midrule
        \multirow{4}{*}{\begin{tabular}[l]{@{}l@{}}TR: Pythia-1B \\ TL: MiniCPM-2B\end{tabular}} 
        & Random & Random & 53.99 \\
        & Best & Random & 54.79 \\
        & Random & Best & 54.51 \\
        & Best & Best & \textbf{55.31} \\
        \midrule
        \multirow{4}{*}{\begin{tabular}[l]{@{}l@{}}TR: MiniCPM-2B \\ TL: Pythia-1B\end{tabular}} 
        & Random & Random & 52.07 \\
        & Best & Random & 53.61 \\
        & Random & Best & 53.73 \\
        & Best & Best & \textbf{55.85} \\
        % \midrule
        % \multirow{4}{*}{\begin{tabular}[l]{@{}l@{}}TR: Pythia-2.8B \\ TL: MiniCPM-2B\end{tabular}} 
        % & Random & Random & 53.32 \\
        % & Best & Random & 53.75 \\
        % & Random & Best & 54.23 \\
        % & Best & Best & \textbf{54.85} \\
        % \midrule
        % \multirow{4}{*}{\begin{tabular}[l]{@{}l@{}}TR: MiniCPM-2B \\ TL: Pythia-2.8B\end{tabular}} 
        % & Random & Random & 54.31 \\
        % & Best & Random & 54.93 \\
        % & Random & Best & 55.97 \\
        % & Best & Best & \textbf{56.38} \\
        \bottomrule
    \end{tabular}
    }
    \caption{Ablation study for model fusion. Results are averaged across 16 datasets. We highlight the highest numbers among fusion with the same models for TR and TL. ``Random'' means that the checkpoint is randomly selected, while ``Best'' means that the checkpoint has the best performance for TR/TL.}
    \label{tab:ckpt-ablation}
\end{table}

\subsection{Results}

As presented in Table~\ref{tab:main-exp}, our proposed method can significantly boost performance compared to the single TR or TL model.
In addition, such an improvement is consistent across various model combinations, demonstrating that our method is widely applicable.
To our surprise, two small models together can even outperform larger models by using this method, despite their total parameters being less than half of the larger ones.
It suggests that our method can effectively fuse the abilities of TR and TL to achieve better ICL performance.

\ignore{In addition, we observe that fusion within the same series of models yields better results.
One possible reason is that they have identical model architecture and pre-training data.
Furthermore, we can see that (TR: Pythia-1B and TL: Pythia-2.8B) performs better than (TR: Pythia-1B and TL: Pythia-1B), demonstrating its great potential to apply to larger language models.}

% 这种方法甚至超过了更大模型的golden setting下的性能，即使总的参数量仅有更大模型的一半不到。
% 我们发现这种融合策略可以在不同系列的模型中实现，即使它们在预训练阶段的相关设置(\eg 模型架构和预训练参数)不完全相同，这表明了方法的有效性和鲁棒性。
% In addition，我们可以发现，在相同系列模型之间融合的效果更好，这可能是因为它们拥有相同的模型架构和预训练数据。
% Furthermore, 对于相同系列的模型来说，增加参与融合模型的规模可以帮助提升性能。

Furthermore, to verify the effectiveness of each component in our method, we conduct the ablation study.
We consider substituting the best checkpoints with random ones or setting the weights of TR and TL to the same, respectively.
As shown in Table~\ref{tab:main-exp} and \ref{tab:ckpt-ablation}, removing any design would lead to a decrease in performance.
It demonstrates the effectiveness of all the components of our approach.
In addition, the selection of checkpoints with the best sub-ability seems to be more important, which yields a larger performance drop after being removed.
Checkpoints with the best sub-ability are more diverse in their predictions, which is important for successful fusion.

\ignore{To illustrate the impact of different checkpoints for fusion, we randomly select three other checkpoints from the pre-training process for replacement and report their average accuracy as the final result.
Table~\ref{tab:ckpt-ablation} shows the impact of different checkpoints. 

The results demonstrate that even when not selecting the most capable checkpoints from the pre-training process, it can still perform better than larger language models with the golden setting in Table~\ref{tab:main-exp}.
This indicates that the adaptive fusion method can effectively combine the capabilities of TR and TL.
Additionally, when checkpoints with stronger TR or TL capabilities are used for fusion, the performance can be further improved.}

% 为了证明我们选取参与融合的两个能力最强模型的必要性，我们随机选取了3个预训练过程中的其它checkpoint分别进行替换，并记录下他们的平均值作为最后的结果。
% Table~\ref{tab:ckpt-ablation} delineates the 参与融合的 different checkpoints 对于实验结果的影响。
% 实验结果证明了（需不需要比2好？）
% 即使我们不选取预训练过程中能力最强的checkpoint，它依然可以超过更大模型的golden setting in Table~\ref{}, 这表明了adaptive fusion method可以充分结合模型TR和TL的能力。
% 当选用TR或TL能力更强的模型参与融合，性能能得到进一步的提升。

% 挑四个数据集
% 四组（1）/（3）
% 1 （3个2）
% 3 （1，1，1+1））（相同模型，相同系列，都不同）
\section{Related Work}
\label{sec-related_work}

Our work is closely related to the studies on the mechanisms of ICL and model fusion.

\paratitle{The Mechanism of ICL.}
Existing work primarily explores the mechanisms of ICL from the pre-training and inference stages of LLMs.
% ~\citet{} demonstrate that small models can acquire ICL abilities by continual pre-training or fine-tuning on specially designed training tasks.
Some work discusses how ICL emerges from pre-training by conducting analysis on pre-training factors like data~\cite{chan2022data, reddy2023mechanistic} and optimization~\cite{singh2024transient, anand2024dual}.
% Some studies demonstrate that the design of pre-training tasks~\cite{} and pre-training data~\cite{} can significantly impact the ICL abilities.
% Other studies explore the influence of pre-training data on ICL abilities, including diversity~\cite{}, long-range dependencies~\cite{}, and token frequency~\cite{}.
Other work~\cite{Disentangle-ACL-2023, Rethinking-2022-EMNLP, dai2023can} studies the operating mechanism of ICL at inference time.
Researchers empirically find two main abilities in ICL: task recognition~(TR) for recognizing the task and utilizing pre-trained priors of LLMs~\cite{Rethinking-2022-EMNLP} and task learning~(TL) for learning from demonstrations~\cite{dai2023can}.
% During inference, as ICL does not involve explicit learning processes, LLMs use examples with two abilities: task recognition~(TR) and task learning~(TL).
% For TR, LLMs can learn and encode hidden variables from examples to form latent vectors~\cite{}, which is then used to trigger the corresponding task recognition process~\cite{}.
% On the other hand, for TL, LLMs can learn new tasks unseen in the pre-training stage only through demonstrations.
% This process is often analyzed from the perspective of gradient descent and considered as implicit fine-tuning~\cite{}.
% Additionally, as discussed in~\cite{}, LLMs exhibit the abilities of both TR and TL.
% They find that smaller models can exhibit TR abilities, while larger models can exhibit strong TL abilities.
In this paper, we explore how TR and TL affect the emergence of ICL.
By examining the pre-training dynamics of LLMs, we demonstrate a strong correlation between the emergence of ICL and the competition between TR and TL.

% After pre-training, LLMs can exhibit intriguing ICL capability without updating parameters.
% Previous work 主要从预训练和推理两个方面探究ICL的内在机制。

% ~\citet{} 通过在专门的训练任务进行继续预训练或持续微调，小模型也能获得上下文学习能力。以此证明了预训练任务的设计对于ICL能力的习得具有重要的影响。
% 还有一些工作探究预训练数据的选择对于ICL能力的影响，这包括预训练数据的多样性~\cite{}，长程依赖关系~\cite{}和token出现频率~\cite{}。
% 在推理阶段，由于上下文学习不涉及显式的学习过程或参数更新，大语言模型使用示例数据的方式主要分为两种范式，包括任务识别和任务学习。
% 在任务识别范式中，大语言模型具备从给定示例中学习并编码这些隐变量的能力~\cite{}, 然后根据这个任务隐向量的指导，在接收到新的输入时自动触发相应的任务识别过程~\cite{}.
% 另一方面，在任务学习范式中，大语言模型具备通过示例数据学习预训练阶段未涉及的新任务的能力。
% 它们一般从梯度下降的角度来分析上下文学习的机理，并将其视为一种隐式的微调过程~\cite{}。
% In addition, 还有一些工作任务ICL是两种范式共同作用的结果~\cite{}，它们发现规模较小的模型已经能展现出较强的任务识别能力，而较大规模的模型才能展现出较强的任务学习能力。

% In this paper, 我们从ICL学习的两种范式中探究预训练的动态。
% 通过定量的实验分析，我们发现两种能力在预训练过程中存在明显的竞争关系， offers new insights and demonstrates the potential for leveraging this understanding to improve the effectiveness and interpretability of ICL.

\paratitle{Model Fusion.}
Model fusion aims to enhance performance by combining the strengths of multiple models~\cite{li2023deep}.
One line of work aims to reduce the difference among different models from perspectives like mode connectivity~\cite{nagarajan2019uniform} and alignment~\cite{tatro2020optimizing}.
Another line of work studies how to leverage the diversity among models through techniques like weight average~\cite{wang2019federated} and ensemble learning~\cite{sagi2018ensemble}.
% Mode Connectivity~\cite{} and Alignment~\cite{} aim to bring the solutions closer to obtain better original conditions of average. 
% However, these methods require significant computational resources when applied to LLMs. 
% Weight Averaging~\cite{} combines several models into a single one in the parameter space in the parameter space without additional computational resources. 
% The most common way to fuse models is ensemble learning~\cite{}, which improves performance and robustness by combining outputs of several different models. 
In this paper, we propose adaptive ensemble learning to fuse checkpoints proficient in TR and TL and achieve better ICL performance.
% This approach achieves remarkable performance compared to a larger model with more than twice the parameters.

% 它们主要有四种方法：Mode Connecticity，Alignment，Weight Averaging and Ensemble Learning.
% Mode Connectivity~\cite{} refers to connecting in weight space by a path (connector) with no obstacles。
% Alignment~\cite{} involves matching units of multiple models to facilitate fusion
% 但是这两种方法应用到在大模型上需要负担很大的计算开销.
% Weight Averaging~\cite{} fuse several parent networks into a single network without additional computational resources.
% 但是直接对模型内部参数进行融合会使得破坏模型的内部结构，resulting in performance degradation.
% Ensemble Learning~\cite{} improve the prediction performance and robustness by combining outputs of several different models.
% In this paper, 我们在推理阶段采用简单而有效的集成学习方法， 融合两个较小的模型，充分释放ICL的两种能力.
\section{Conclusion}
\label{sec-conclusion}

In this paper, we presented the first study of the competitive relationship of TR and TL abilities, and quantified their effect on the emergence of ICL. 
%take the first step towards exploring the effect of dual abilities in ICL (\ie task recognition and task learning) on its emergence.
With specially designed metrics, we found that the competition between dual abilities widely exists in existing LLMs, and the competition intensity is negatively correlated with the ICL performance.
Then, we conducted a detailed analysis of several pre-training factors (\ie model size, dataset size, and data curriculum) to demonstrate possible ways to regulate the competition.
Furthermore, we proposed a simple yet effective method to better integrate dual abilities at inference time.
Through adaptive ensemble learning, the performance of ICL can be significantly boosted, enabling two small models to outperform a larger one with more than twice the parameters.
Overall, our work provides new insights and approaches to study and understand the underlying mechanism of ICL, which is worth deep exploration for improving the capacity of LLMs. 
%Overall, our work opens up new directions for understanding the mechanism of ICL and exploiting its full potential.
\section{Limitations}
\label{sec-limitations}

Although our study provides valuable insights into the dual abilities of ICL and its emergence, several limitations should be noted.
First, our research focuses on classification tasks since they can be easily adapted for the three evaluation settings of ICL.
Other types of tasks are left for future work.
Second, our investigation is confined to conventional ICL paradigms and does not explore alternative paradigms such as chain-of-thought prompting~(CoT).
Third, due to computational constraints, our study mainly considers LLMs with up to 12 billion parameters.
Replicating our study with larger-scale LLMs could provide further insights and validate the robustness of our findings across different model sizes.

% \section*{Acknowledgments}

\bibliography{newbib}

\begin{thebibliography}{37}
\providecommand{\natexlab}[1]{#1}

\bibitem[{Anand et~al.(2024)Anand, Lepori, Merullo, and Pavlick}]{anand2024dual}
Suraj Anand, Michael~A Lepori, Jack Merullo, and Ellie Pavlick. 2024.
\newblock Dual process learning: Controlling use of in-context vs. in-weights strategies with weight forgetting.
\newblock \emph{arXiv preprint arXiv:2406.00053}.

\bibitem[{Basile et~al.(2019)Basile, Bosco, Fersini, Nozza, Patti, Pardo, Rosso, and Sanguinetti}]{tweet-2019-dataset}
Valerio Basile, Cristina Bosco, Elisabetta Fersini, Debora Nozza, Viviana Patti, Francisco Manuel~Rangel Pardo, Paolo Rosso, and Manuela Sanguinetti. 2019.
\newblock Semeval-2019 task 5: Multilingual detection of hate speech against immigrants and women in twitter.
\newblock In \emph{SemEval@NAACL-HLT}, pages 54--63. Association for Computational Linguistics.

\bibitem[{Biderman et~al.(2023)Biderman, Schoelkopf, Anthony, Bradley, O'Brien, Hallahan, Khan, Purohit, Prashanth, Raff, Skowron, Sutawika, and van~der Wal}]{Pythia-model}
Stella Biderman, Hailey Schoelkopf, Quentin~Gregory Anthony, Herbie Bradley, Kyle O'Brien, Eric Hallahan, Mohammad~Aflah Khan, Shivanshu Purohit, USVSN~Sai Prashanth, Edward Raff, Aviya Skowron, Lintang Sutawika, and Oskar van~der Wal. 2023.
\newblock Pythia: {A} suite for analyzing large language models across training and scaling.
\newblock In \emph{{ICML}}, volume 202 of \emph{Proceedings of Machine Learning Research}, pages 2397--2430. {PMLR}.

\bibitem[{Bowman et~al.(2015)Bowman, Angeli, Potts, and Manning}]{snli-dataset}
Samuel~R. Bowman, Gabor Angeli, Christopher Potts, and Christopher~D. Manning. 2015.
\newblock A large annotated corpus for learning natural language inference.
\newblock In \emph{{EMNLP}}, pages 632--642. The Association for Computational Linguistics.

\bibitem[{Brown et~al.(2020)Brown, Mann, Ryder, Subbiah, Kaplan, Dhariwal, Neelakantan, Shyam, Sastry, Askell et~al.}]{brown2020language}
Tom Brown, Benjamin Mann, Nick Ryder, Melanie Subbiah, Jared~D Kaplan, Prafulla Dhariwal, Arvind Neelakantan, Pranav Shyam, Girish Sastry, Amanda Askell, et~al. 2020.
\newblock Language models are few-shot learners.
\newblock \emph{Advances in neural information processing systems}, 33:1877--1901.

\bibitem[{Chan et~al.(2022)Chan, Santoro, Lampinen, Wang, Singh, Richemond, McClelland, and Hill}]{chan2022data}
Stephanie Chan, Adam Santoro, Andrew Lampinen, Jane Wang, Aaditya Singh, Pierre Richemond, James McClelland, and Felix Hill. 2022.
\newblock Data distributional properties drive emergent in-context learning in transformers.
\newblock \emph{Advances in Neural Information Processing Systems}, 35:18878--18891.

\bibitem[{Dai et~al.(2023)Dai, Sun, Dong, Hao, Ma, Sui, and Wei}]{dai2023can}
Damai Dai, Yutao Sun, Li~Dong, Yaru Hao, Shuming Ma, Zhifang Sui, and Furu Wei. 2023.
\newblock Why can gpt learn in-context? language models secretly perform gradient descent as meta-optimizers.
\newblock In \emph{Findings of the Association for Computational Linguistics: ACL 2023}, pages 4005--4019.

\bibitem[{Dolan and Brockett(2005)}]{mrpc-dataset}
William~B. Dolan and Chris Brockett. 2005.
\newblock Automatically constructing a corpus of sentential paraphrases.
\newblock In \emph{IWP@IJCNLP}. Asian Federation of Natural Language Processing.

\bibitem[{Dong et~al.(2022)Dong, Li, Dai, Zheng, Wu, Chang, Sun, Xu, and Sui}]{dong2022survey}
Qingxiu Dong, Lei Li, Damai Dai, Ce~Zheng, Zhiyong Wu, Baobao Chang, Xu~Sun, Jingjing Xu, and Zhifang Sui. 2022.
\newblock A survey on in-context learning.
\newblock \emph{arXiv preprint arXiv:2301.00234}.

\bibitem[{Groeneveld et~al.(2024)Groeneveld, Beltagy, Walsh, Bhagia, Kinney, Tafjord, Jha, Ivison, Magnusson, Wang, Arora, Atkinson, Authur, Chandu, Cohan, Dumas, Elazar, Gu, Hessel, Khot, Merrill, Morrison, Muennighoff, Naik, Nam, Peters, Pyatkin, Ravichander, Schwenk, Shah, Smith, Strubell, Subramani, Wortsman, Dasigi, Lambert, Richardson, Zettlemoyer, Dodge, Lo, Soldaini, Smith, and Hajishirzi}]{Olmo-model}
Dirk Groeneveld, Iz~Beltagy, Pete Walsh, Akshita Bhagia, Rodney Kinney, Oyvind Tafjord, Ananya~Harsh Jha, Hamish Ivison, Ian Magnusson, Yizhong Wang, Shane Arora, David Atkinson, Russell Authur, Khyathi~Raghavi Chandu, Arman Cohan, Jennifer Dumas, Yanai Elazar, Yuling Gu, Jack Hessel, Tushar Khot, William Merrill, Jacob Morrison, Niklas Muennighoff, Aakanksha Naik, Crystal Nam, Matthew~E. Peters, Valentina Pyatkin, Abhilasha Ravichander, Dustin Schwenk, Saurabh Shah, Will Smith, Emma Strubell, Nishant Subramani, Mitchell Wortsman, Pradeep Dasigi, Nathan Lambert, Kyle Richardson, Luke Zettlemoyer, Jesse Dodge, Kyle Lo, Luca Soldaini, Noah~A. Smith, and Hannaneh Hajishirzi. 2024.
\newblock Olmo: Accelerating the science of language models.
\newblock \emph{CoRR}, abs/2402.00838.

\bibitem[{Hu et~al.(2024)Hu, Tu, Han, He, Cui, Long, Zheng, Fang, Huang, Zhao, Zhang, Thai, Zhang, Wang, Yao, Zhao, Zhou, Cai, Zhai, Ding, Jia, Zeng, Li, Liu, and Sun}]{Minicpm-model}
Shengding Hu, Yuge Tu, Xu~Han, Chaoqun He, Ganqu Cui, Xiang Long, Zhi Zheng, Yewei Fang, Yuxiang Huang, Weilin Zhao, Xinrong Zhang, Zhen~Leng Thai, Kai Zhang, Chongyi Wang, Yuan Yao, Chenyang Zhao, Jie Zhou, Jie Cai, Zhongwu Zhai, Ning Ding, Chao Jia, Guoyang Zeng, Dahai Li, Zhiyuan Liu, and Maosong Sun. 2024.
\newblock Minicpm: Unveiling the potential of small language models with scalable training strategies.
\newblock \emph{CoRR}, abs/2404.06395.

\bibitem[{Kaplan et~al.(2020)Kaplan, McCandlish, Henighan, Brown, Chess, Child, Gray, Radford, Wu, and Amodei}]{kaplan2020scaling}
Jared Kaplan, Sam McCandlish, Tom Henighan, Tom~B Brown, Benjamin Chess, Rewon Child, Scott Gray, Alec Radford, Jeffrey Wu, and Dario Amodei. 2020.
\newblock Scaling laws for neural language models.
\newblock \emph{arXiv preprint arXiv:2001.08361}.

\bibitem[{Levesque et~al.(2012)Levesque, Davis, and Morgenstern}]{wnli-dataset}
Hector~J. Levesque, Ernest Davis, and Leora Morgenstern. 2012.
\newblock The winograd schema challenge.
\newblock In \emph{{KR}}. {AAAI} Press.

\bibitem[{Li et~al.(2023{\natexlab{a}})Li, Zi, Muennighoff, Kocetkov, Mou, Marone, Akiki, Jia, Chim, Liu et~al.}]{li2023starcoder}
Raymond Li, Yangtian Zi, Niklas Muennighoff, Denis Kocetkov, Chenghao Mou, Marc Marone, Christopher Akiki, LI~Jia, Jenny Chim, Qian Liu, et~al. 2023{\natexlab{a}}.
\newblock Starcoder: may the source be with you!
\newblock \emph{Transactions on Machine Learning Research}.

\bibitem[{Li et~al.(2023{\natexlab{b}})Li, Peng, Zhang, Ding, Hu, and Shen}]{li2023deep}
Weishi Li, Yong Peng, Miao Zhang, Liang Ding, Han Hu, and Li~Shen. 2023{\natexlab{b}}.
\newblock Deep model fusion: A survey.
\newblock \emph{arXiv preprint arXiv:2309.15698}.

\bibitem[{Lin et~al.(2023)Lin, Ravichander, Lu, Dziri, Sclar, Chandu, Bhagavatula, and Choi}]{lin2023unlocking}
Bill~Yuchen Lin, Abhilasha Ravichander, Ximing Lu, Nouha Dziri, Melanie Sclar, Khyathi Chandu, Chandra Bhagavatula, and Yejin Choi. 2023.
\newblock The unlocking spell on base llms: Rethinking alignment via in-context learning.
\newblock \emph{arXiv preprint arXiv:2312.01552}.

\bibitem[{Lin and Lee(2024)}]{lin2024dual}
Ziqian Lin and Kangwook Lee. 2024.
\newblock Dual operating modes of in-context learning.
\newblock \emph{arXiv preprint arXiv:2402.18819}.

\bibitem[{Liu et~al.(2023)Liu, Qiao, Neiswanger, Wang, Tan, Tao, Li, Wang, Sun, Pangarkar, Fan, Gu, Miller, Zhuang, He, Li, Koto, Tang, Ranjan, Shen, Ren, Iriondo, Mu, Hu, Schulze, Nakov, Baldwin, and Xing}]{LLM360-model}
Zhengzhong Liu, Aurick Qiao, Willie Neiswanger, Hongyi Wang, Bowen Tan, Tianhua Tao, Junbo Li, Yuqi Wang, Suqi Sun, Omkar Pangarkar, Richard Fan, Yi~Gu, Victor Miller, Yonghao Zhuang, Guowei He, Haonan Li, Fajri Koto, Liping Tang, Nikhil Ranjan, Zhiqiang Shen, Xuguang Ren, Roberto Iriondo, Cun Mu, Zhiting Hu, Mark Schulze, Preslav Nakov, Tim Baldwin, and Eric~P. Xing. 2023.
\newblock {LLM360:} towards fully transparent open-source llms.
\newblock \emph{CoRR}, abs/2312.06550.

\bibitem[{Malo et~al.(2014)Malo, Sinha, Korhonen, Wallenius, and Takala}]{financial_phrasebank-dataset}
Pekka Malo, Ankur Sinha, Pekka~J. Korhonen, Jyrki Wallenius, and Pyry Takala. 2014.
\newblock Good debt or bad debt: Detecting semantic orientations in economic texts.
\newblock \emph{J. Assoc. Inf. Sci. Technol.}, 65(4):782--796.

\bibitem[{Marelli et~al.(2014)Marelli, Menini, Baroni, Bentivogli, Bernardi, and Zamparelli}]{sick-dataset}
Marco Marelli, Stefano Menini, Marco Baroni, Luisa Bentivogli, Raffaella Bernardi, and Roberto Zamparelli. 2014.
\newblock A {SICK} cure for the evaluation of compositional distributional semantic models.
\newblock In \emph{{LREC}}, pages 216--223. European Language Resources Association {(ELRA)}.

\bibitem[{Min et~al.(2022)Min, Lyu, Holtzman, Artetxe, Lewis, Hajishirzi, and Zettlemoyer}]{Rethinking-2022-EMNLP}
Sewon Min, Xinxi Lyu, Ari Holtzman, Mikel Artetxe, Mike Lewis, Hannaneh Hajishirzi, and Luke Zettlemoyer. 2022.
\newblock Rethinking the role of demonstrations: What makes in-context learning work?
\newblock In \emph{{EMNLP}}, pages 11048--11064. Association for Computational Linguistics.

\bibitem[{Mohammad et~al.(2018)Mohammad, Bravo{-}Marquez, Salameh, and Kiritchenko}]{tweet-2018-dataset}
Saif~M. Mohammad, Felipe Bravo{-}Marquez, Mohammad Salameh, and Svetlana Kiritchenko. 2018.
\newblock Semeval-2018 task 1: Affect in tweets.
\newblock In \emph{SemEval@NAACL-HLT}, pages 1--17. Association for Computational Linguistics.

\bibitem[{Mollas et~al.(2020)Mollas, Chrysopoulou, Karlos, and Tsoumakas}]{ethos-dataset}
Ioannis Mollas, Zoe Chrysopoulou, Stamatis Karlos, and Grigorios Tsoumakas. 2020.
\newblock {ETHOS:} an online hate speech detection dataset.
\newblock \emph{CoRR}, abs/2006.08328.

\bibitem[{Nagarajan and Kolter(2019)}]{nagarajan2019uniform}
Vaishnavh Nagarajan and J~Zico Kolter. 2019.
\newblock Uniform convergence may be unable to explain generalization in deep learning.
\newblock \emph{Advances in Neural Information Processing Systems}, 32.

\bibitem[{Pan et~al.(2023)Pan, Gao, Chen, and Chen}]{Disentangle-ACL-2023}
Jane Pan, Tianyu Gao, Howard Chen, and Danqi Chen. 2023.
\newblock What in-context learning "learns" in-context: Disentangling task recognition and task learning.
\newblock In \emph{{ACL} (Findings)}, pages 8298--8319. Association for Computational Linguistics.

\bibitem[{Reddy(2023)}]{reddy2023mechanistic}
Gautam Reddy. 2023.
\newblock The mechanistic basis of data dependence and abrupt learning in an in-context classification task.
\newblock In \emph{The Twelfth International Conference on Learning Representations}.

\bibitem[{Sagi and Rokach(2018)}]{sagi2018ensemble}
Omer Sagi and Lior Rokach. 2018.
\newblock Ensemble learning: A survey.
\newblock \emph{Wiley interdisciplinary reviews: data mining and knowledge discovery}, 8(4):e1249.

\bibitem[{Saravia et~al.(2018)Saravia, Liu, Huang, Wu, and Chen}]{emotion-dataset}
Elvis Saravia, Hsien{-}Chi~Toby Liu, Yen{-}Hao Huang, Junlin Wu, and Yi{-}Shin Chen. 2018.
\newblock {CARER:} contextualized affect representations for emotion recognition.
\newblock In \emph{{EMNLP}}, pages 3687--3697. Association for Computational Linguistics.

\bibitem[{Sheng and Uthus(2020)}]{poem-dataset}
Emily Sheng and David~C. Uthus. 2020.
\newblock Investigating societal biases in a poetry composition system.
\newblock \emph{CoRR}, abs/2011.02686.

\bibitem[{Singh et~al.(2024)Singh, Chan, Moskovitz, Grant, Saxe, and Hill}]{singh2024transient}
Aaditya Singh, Stephanie Chan, Ted Moskovitz, Erin Grant, Andrew Saxe, and Felix Hill. 2024.
\newblock The transient nature of emergent in-context learning in transformers.
\newblock \emph{Advances in Neural Information Processing Systems}, 36.

\bibitem[{Soboleva et~al.(2023)Soboleva, Al-Khateeb, Myers, Steeves, Hestness, and Dey}]{cerebras2023slimpajama}
Daria Soboleva, Faisal Al-Khateeb, Robert Myers, Jacob~R Steeves, Joel Hestness, and Nolan Dey. 2023.
\newblock \href {https://huggingface.co/datasets/cerebras/SlimPajama-627B} {{SlimPajama: A 627B token cleaned and deduplicated version of RedPajama}}.
\newblock \url{https://www.cerebras.net/blog/slimpajama-a-627b-token-cleaned-and-deduplicated-version-of-redpajama}.

\bibitem[{Socher et~al.(2013)Socher, Perelygin, Wu, Chuang, Manning, Ng, and Potts}]{sst2-dataset}
Richard Socher, Alex Perelygin, Jean Wu, Jason Chuang, Christopher~D. Manning, Andrew~Y. Ng, and Christopher Potts. 2013.
\newblock Recursive deep models for semantic compositionality over a sentiment treebank.
\newblock In \emph{{EMNLP}}, pages 1631--1642. {ACL}.

\bibitem[{Tatro et~al.(2020)Tatro, Chen, Das, Melnyk, Sattigeri, and Lai}]{tatro2020optimizing}
Norman Tatro, Pin-Yu Chen, Payel Das, Igor Melnyk, Prasanna Sattigeri, and Rongjie Lai. 2020.
\newblock Optimizing mode connectivity via neuron alignment.
\newblock \emph{Advances in Neural Information Processing Systems}, 33:15300--15311.

\bibitem[{Voorhees and Tice(2000)}]{trec-dataset}
Ellen~M. Voorhees and Dawn~M. Tice. 2000.
\newblock Building a question answering test collection.
\newblock In \emph{{SIGIR}}, pages 200--207. {ACM}.

\bibitem[{Wang et~al.(2019)Wang, Yurochkin, Sun, Papailiopoulos, and Khazaeni}]{wang2019federated}
Hongyi Wang, Mikhail Yurochkin, Yuekai Sun, Dimitris Papailiopoulos, and Yasaman Khazaeni. 2019.
\newblock Federated learning with matched averaging.
\newblock In \emph{International Conference on Learning Representations}.

\bibitem[{Wei et~al.(2023)Wei, Wei, Tay, Tran, Webson, Lu, Chen, Liu, Huang, Zhou et~al.}]{wei2023larger}
Jerry Wei, Jason Wei, Yi~Tay, Dustin Tran, Albert Webson, Yifeng Lu, Xinyun Chen, Hanxiao Liu, Da~Huang, Denny Zhou, et~al. 2023.
\newblock Larger language models do in-context learning differently.
\newblock \emph{arXiv preprint arXiv:2303.03846}.

\bibitem[{Yang et~al.(2023)Yang, Xiao, Wang, Zhang, Bian, Yin, Lv, Pan, Wang, Yan, Yang, Deng, Wang, Liu, Ai, Dong, Zhao, Xu, Sun, Zhang, Liu, Ji, Xie, Dai, Fang, Su, Song, Liu, Ru, Ma, Wang, Liu, Lin, Nie, Guo, Sun, Zhang, Li, Li, Cheng, Chen, Zeng, Wang, Chen, Men, Yu, Pan, Shen, Wang, Li, Jiang, Gao, Zhang, Zhou, and Wu}]{Baichuan2-model}
Aiyuan Yang, Bin Xiao, Bingning Wang, Borong Zhang, Ce~Bian, Chao Yin, Chenxu Lv, Da~Pan, Dian Wang, Dong Yan, Fan Yang, Fei Deng, Feng Wang, Feng Liu, Guangwei Ai, Guosheng Dong, Haizhou Zhao, Hang Xu, Haoze Sun, Hongda Zhang, Hui Liu, Jiaming Ji, Jian Xie, Juntao Dai, Kun Fang, Lei Su, Liang Song, Lifeng Liu, Liyun Ru, Luyao Ma, Mang Wang, Mickel Liu, MingAn Lin, Nuolan Nie, Peidong Guo, Ruiyang Sun, Tao Zhang, Tianpeng Li, Tianyu Li, Wei Cheng, Weipeng Chen, Xiangrong Zeng, Xiaochuan Wang, Xiaoxi Chen, Xin Men, Xin Yu, Xuehai Pan, Yanjun Shen, Yiding Wang, Yiyu Li, Youxin Jiang, Yuchen Gao, Yupeng Zhang, Zenan Zhou, and Zhiying Wu. 2023.
\newblock Baichuan 2: Open large-scale language models.
\newblock \emph{CoRR}, abs/2309.10305.

\end{thebibliography}
\clearpage
\appendix

\section{Tasks and Datasets}
\label{app:detailed-exp-settings}

% \subsection{Datasets}

We conduct experiments on four types of tasks: Sentiment Analysis, Topic/Stance Classification, Toxicity Detection, and Natural Language Inference/Paraphrase Detection.
For \textbf{Sentiment Analysis}, we use datasets including SST-2~\cite{sst2-dataset}, financial\_phrasebank~\cite{financial_phrasebank-dataset}, emotion~\cite{emotion-dataset}, and poem\_sentiment~\cite{poem-dataset}.
For \textbf{Topic/Stance Classification}, we utilize TREC~\cite{trec-dataset}, tweet\_eval\_atheist, and tweet\_eval\_feminist~\cite{tweet-2018-dataset, tweet-2019-dataset}.
For \textbf{Toxicity Detection}, we include tweet\_eval\_hate, ethos\_race, ethos\_gender, ethos\_national\_origin, and ethos\_religion~\cite{ethos-dataset}.
For \textbf{Natural Language Inference/Paraphrase Detection}, we employ SICK~\cite{sick-dataset}, SNLI~\cite{snli-dataset}, WNLI~\cite{wnli-dataset}, and MRPC~\cite{mrpc-dataset}.

We follow ~\citet{Rethinking-2022-EMNLP} to select samples from the training set as demonstrations.
Additionally, we randomly sample 300 examples as the development set for validation in Section~\ref{sec-method} and another 1000 examples as the test set for evaluation in all experiments from the development set.

% 我们在四个类型的任务上跑实验：Sentiment Analysis, Topic/stance classification, Toxicity detection, and Natural language inference/paraphrase detection.
% Sentiment Analysis includes SST-2~\cite{}, financial_phrasebank~\cite{}, emotion~\cite{}, and poem_sentiment~\cite{}.
% Topic/stance classification includes TREC~\cite{}, tweet_eval_atheist, and tweet_eval_feminist~\cite{}. 
% Toxicity detection includes tweet_eval_hate, ethos_race, ethos_gender, ethos_national_origin, and ethos_religion~\cite{}.
% Natural language inference/paraphrase detection includes SICK~\cite{}, SNLI~\cite{}, WNLI~\cite{}, and MRPC~\cite{}.
% 我们follow~\cite{rethinking} 从训练集中选取sample作为demonstrations，并且从验证集中随机选取300个示例作为验证集，选取另外的1000个示例作为测试集。

% 示例

\section{More Experiments}
\label{app:exp}

\subsection{The Number of Intermediate Checkpoints}
\label{app-sub:more-ckpt}

\begin{table}[t]
\centering
    % \small
    \resizebox{\columnwidth}{!}{%    
    \begin{tabular}{c|c|c|c}
    \toprule
    \begin{tabular}[c]{@{}c@{}}\textbf{\# Intermediate checkpoint}\end{tabular} & 8 & 16 & 32 \\ 
    \midrule
    Pythia-6.9B & 12.50 & 37.50 & 53.12 \\ 
    \midrule
    OLMo-7B & 50.00 & 37.50 & 43.75 \\ 
    \midrule
    MiniCPM-2B & 62.50 & 56.25 & 37.50 \\ 
    \bottomrule
    \end{tabular}
    }
    \caption{Different numbers of intermediate checkpoints}
    \label{tab:app-ckpt}
\end{table}

In the paper, we use 16 checkpoints in addition to the final one.
In this part, we conduct experiments using different numbers of checkpoints (\ie 8 and 32).
We report the average competition ratio across 16 datasets and 5 random seeds.
Table~\ref{tab:app-ckpt} shows that the number of checkpoints does not affect the experimental results. 
They consistently demonstrate that there is a competitive relationship between TR and TL during the pre-training process.

% 在正文中，我们采用了16个checkpoints in addition to the final one.
% 在这个部分，我们使用使用其他数量的checkpoints (\ie 8 and 32)开展实验。
% We report the average existence of competition across 16 datasets and 5 random seeds. 
% 实验结果表明，选取的checkpoint数量并不影响实验结果，他们都能说明模型在预训练过程中dual abilities存在竞争关系。

% minicpm, pythia-6.9b, olmo

\subsection{The Numbers of Examples in Demonstration}
\label{app-sub:more-demo}

\begin{table}[t]
\centering
    \small
    % \resizebox{\columnwidth}{!}{%    
    \begin{tabular}{c|c|c|c}
    \toprule
    \begin{tabular}[c]{@{}c@{}}\textbf{\# Examples}\end{tabular} & 4 & 8 & 16 \\ 
    \midrule
    Pythia-6.9B & 18.75 & 25.00 & 37.50 \\ 
    \midrule
    OLMo-7B & 5.00 & 37.50 & 37.50 \\ 
    \midrule
    MiniCPM-2B & 43.75 & 31.25 & 56.25 \\ 
    \bottomrule
    \end{tabular}
    % }
    \caption{Different numbers of examples in demonstration}
    \label{tab:app-demo}
\end{table}

In the paper, we use 16 randomly sampled examples as demonstrations.
To explore the impact of the number of examples, we report the average competition ratio with other numbers (\ie 4 and 8) of demonstrations.
As presented in Table~\ref{tab:app-demo}, it can be observed that the number of examples does not affect the competitive relationship during the pre-training process.

% 在正文中，我们使用了16 randomly sampled examples as demonstrations.
% 为了探究示例数量影响，我们使用其他数量 (\ie 4 and 8) 的demonstrations 开展实验。
% 实验结果表明，示例数量并不会影响模型在预训练过程中的竞争关系

\subsection{The Type of Abstract Labels}
\label{app-sub:more-abstract}

\begin{table}[t]
\centering
    \small
    % \resizebox{\columnwidth}{!}{%    
    \begin{tabular}{c|c|c|c}
    \toprule
    \begin{tabular}[c]{@{}c@{}}\textbf{Abstract} \\ \textbf{labels}\end{tabular} & \begin{tabular}[c]{@{}c@{}}Symbols\end{tabular} & \begin{tabular}[c]{@{}c@{}}Numbers\end{tabular} & \begin{tabular}[c]{@{}c@{}}Letters\end{tabular} \\ 
    \midrule
    Pythia-6.9B & 37.50 & 18.75 & 37.50 \\ 
    \midrule
    OLMo-7B & 37.50 & 50.00 & 43.75 \\ 
    \midrule
    MiniCPM-2B & 56.25 & 43.75 & 50.00 \\ 
    \bottomrule
    \end{tabular}
    % }
    \caption{Different types of abstract labels.}
    \label{tab:app-abstract}
\end{table}

In the paper, we utilize symbols in the abstract setting.
In this part, we follow~\cite{Disentangle-ACL-2023} to use other types of semantically unrelated labels (\ie numbers and letters).
Table~\ref{tab:app-abstract} shows the average competition ratio by using different labels.
It indicates that, regardless of the choice of semantically unrelated labels, the conclusions are consistent with the abstract symbols.

% 在正文中，我们使用了abstract symbol的方式作为abstract setting.
% In this part, 我们follow~\cite{} 使用其他类型的semantically unrelated labels (\ie numbers and letters)。
% 实验结果表明，无论选择何种语义无关的labels，结论与正文中选取的symbols一致。

% \section{More Results}
% \label{app-sub:more-results}

% \tablename~\ref{tab:ckpt-ablation} shows the results of the ablation study for our proposed method in Section~\ref{sec-method}.

% \figurename~\ref{fig:performance-amber} shows the performance of Amber-7B for ICL, TR, and TL.

% \figurename~\ref{fig:dynamic-amber} shows the performance of ICL and the evolution of competition during the pre-training of Amber-7B.

% \figurename~\ref{fig:dataset-size-small-model} shows the competition evolution of LLMs trained with different dataset sizes.

% \clearpage

% \begin{figure}[t]
%     \centering
%     \includegraphics[width=\columnwidth]{figures/Amber-Performance.pdf}
%     \caption{The performance of Amber-7B for ICL, TR, and TL.}
% \label{fig:performance-amber}
% \end{figure}

% \begin{figure}[t]
%     \centering
%     \includegraphics[width=\columnwidth]{figures/golden-competitiveness-amber.pdf}
%     \caption{The performance of ICL and the evolution of competition ($R_i$) during the pre-training of Amber-7B.}
% \label{fig:dynamic-amber}
% \end{figure}

% \begin{figure}[t]
%     \centering
%     \includegraphics[width=\linewidth]{figures/small_model_token.pdf}
%     \caption{
%         Competition evolution of LLMs trained with different dataset sizes.
%     }
%     \label{fig:dataset-size-small-model}
% \end{figure}

\end{document}